%% file: main.tex
\documentclass[10pt,twocolumn,letterpaper]{article}

\usepackage[accsupp]{axessibility}  

\usepackage{wacv}              

\input{preamble}

\definecolor{wacvblue}{rgb}{0.21,0.49,0.74}
\usepackage[pagebackref,breaklinks,colorlinks,allcolors=wacvblue]{hyperref}

\def\wacvPaperID{148} 
\def\confName{WACV}
\def\confYear{2026}

\title{MomentMix Augmentation with Length-Aware DETR \\ for Temporally Robust Moment Retrieval}

\author{\textbf{Seojeong Park}$^1$\quad\quad \textbf{Jiho Choi}$^1$\quad\quad \textbf{Kyungjune Baek}$^2$\quad\quad \textbf{Hyunjung Shim}$^1$\\
$^1$Korea Advanced Institute of Science and Technology (KAIST), Republic of Korea \\ $^2$Sejong University, Republic of Korea\\
{\tt\small \{seojeong.park, jihochoi, kateshim\}@kaist.ac.kr,   kyungjune.baek@sejong.ac.kr}\\
}

\begin{document}

\maketitle

\input{sections/00_abstract}

\input{sections/01_introduction}

\input{sections/02_related_work}

\input{sections/03_method}

\input{sections/04_experiments}

\input{sections/05_conclusion}

{
    \small
    \bibliographystyle{ieeenat_fullname}
    \bibliography{main}
}

\input{sections/06_supplementary_materials}

\end{document}

%% file: preamble.tex
\usepackage{threeparttable}        
\usepackage{wrapfig}               
\usepackage{booktabs}
\usepackage{xcolor}                
\usepackage{bbding}                
\usepackage{amsmath}               
\usepackage{graphicx}              
\usepackage{accents}               
\usepackage{multirow}              
\usepackage{mathtools}             
\usepackage{footnote}
\usepackage{tablefootnote}
\usepackage{esvect}                
\usepackage{tabularx}              
\usepackage{subfiles}
\usepackage{colortbl}              
\usepackage{amssymb}               
\usepackage{comment}               
\usepackage{titletoc}              
\usepackage{tocbibind}             
\usepackage{appendix}              
\usepackage{url}

\usepackage{inconsolata}           

\usepackage{lipsum}                
\usepackage{kotex}                 

\definecolor{darkergreen}{RGB}{6, 150, 62}
\definecolor{red2}{RGB}{171, 60, 46}
\newcommand{\yesmark}{\textcolor{darkergreen}{\ding{52}}}
\newcommand{\nomark}{\textcolor{red2}{\ding{56}}}

\definecolor{tealgreen}{rgb}{0.0, 0.51, 0.5}
\newcommand{\gainp}[1]{\textcolor{darkergreen}{$^{\textbf{\texttt{(#1)}}}$}}
\newcommand{\gainm}[1]{\textcolor{red2}{$^{\textbf{\texttt{(#1)}}}$}}

\usepackage{dblfloatfix}

\usepackage{tabularx}
\usepackage{xcolor}              
\usepackage{pifont}

\DeclareMathOperator*{\argmin}{arg\,min}

\newcommand{\MODELNAME}[1]{{LA-DETR}}
\usepackage{dsfont}

%% file: sections/00_abstract.tex
\begin{abstract}
    Video Moment Retrieval (MR) aims to localize moments within a video based on a given natural language query.
    Given the prevalent use of platforms like YouTube for information retrieval, the demand for MR techniques is significantly growing.
    Recent DETR-based models have made notable advances in performance but still struggle with accurately localizing short moments.
    Through data analysis, we identified limited feature diversity in short moments, which motivated the development of MomentMix. MomentMix generates new short-moment samples by employing two augmentation strategies: ForegroundMix and BackgroundMix, each enhancing the ability to understand the query-relevant and irrelevant frames, respectively. Additionally, our analysis of prediction bias revealed that short moments particularly struggle with accurately predicting their center positions and length of moments. To address this, we propose a Length-Aware Decoder, which conditions length through a novel bipartite matching process.
    Our extensive studies demonstrate the efficacy of our length-aware approach, especially in localizing short moments, leading to improved overall performance.
    Our method surpasses state-of-the-art DETR-based methods on benchmark datasets, achieving the highest R1 and mAP on QVHighlights and the highest R1@0.7 on TACoS and Charades-STA (such as a 9.62\% gain in R1@0.7 and an 16.9\% gain in mAP average for QVHighlights). The code is available at \url{https://github.com/sjpark5800/LA-DETR}.
\end{abstract}

%% file: sections/01_introduction.tex
%

\vspace{-2mm}

\section{Introduction}
\label{sec:introduction}


As vast amounts of video content are created and shared on the internet daily~\cite{cheng2013understanding_YouTube}, moment retrieval (MR)~\cite{anne2017localizing,gao2017tall_tall} has gained significant attention to improve user experience and search efficiency. MR identifies the specific moments within a video that best align with a given query. Specifically, this task involves localizing the start and end points in the video relevant to the textual query, offering a more detailed understanding of video content.

\begin{figure}[tb]
    \centering

    \includegraphics[width=1.0\linewidth]{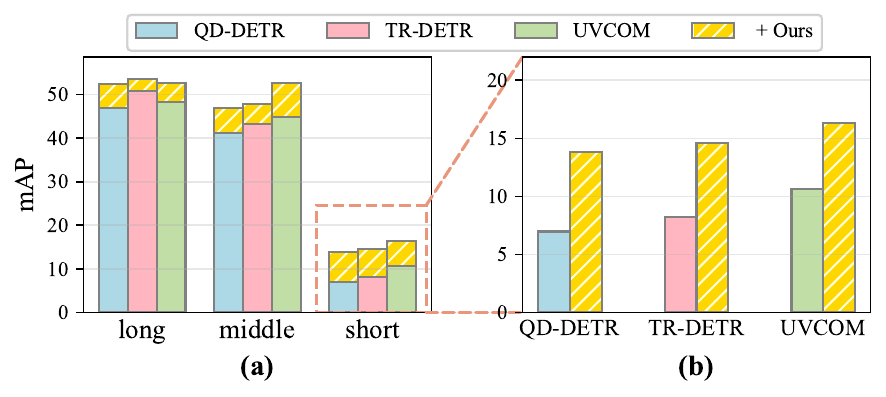}
    \caption{
        {(a) 
        Performance (mAP) of previous moment retrieval (MR) DETR-based methods \cite{moon2023query_QD-DETR,sun2024tr_TR-DETR,xiao2024bridging_UVCOM} on \textsc{QVHighlights} \textit{test} set by the lengths of the moment. Notice that the mAP drops significantly in capturing short-moment, where MR can be best utilized.
        (b) Short moment performance (mAP) comparison.}
    }
    \vspace{-4mm}
    \label{fig:intro_short}
\end{figure}


For the MR task, various methods have been explored, including DETR-based approaches~\cite{lei2021detecting_Moment-DETR,moon2023query_QD-DETR, moon2023correlation_CG-DETR, sun2024tr_TR-DETR, xiao2024bridging_UVCOM}, non-DETR models~\cite{lin2023univtg_Univtg, liu2024r2-tuning, cao2024flashvtg}, and models leveraging the power of LLMs~\cite{ren2024timechat, huang2024vtimellm, jiang2024LLMEPET}. LLM-based approaches achieve strong zero-shot performance but are impractical for real-world applications due to slow inference speeds. Non-DETR methods offer faster inference but rely heavily on complex post-processing, such as Non-Maximum Suppression (NMS), making them sensitive to IoU thresholds. DETR-based models have attracted significant attention by addressing these issues through their fast inference speeds and robust set-prediction framework, enabling accurate predictions without additional post-processing. However, our empirical findings indicate that these DETR-based models suffer from a significant drop in performance when handling short moments as highlighted in \Cref{fig:intro_short}.
For example, UVCOM~\cite{xiao2024bridging_UVCOM} shows an average mAP of 44.90 for middle-length moments (10–30 seconds), 
while achieving only 10.67 for short moments (less than 10 seconds), revealing a substantial gap.


Retrieving short moments within videos is a crucial task because videos often contain a significant amount of redundant or irrelevant information, while essential content is frequently condensed in short moments. This aligns with the importance of MR, where improving the accuracy of short-moment retrieval enables the precise extraction of the most relevant information. 
Such improvements can significantly reduce the time and effort required for video exploration.
For example, highlights in sports and news, as well as key scenes in movies and dramas, often involve short moments. This emphasizes the importance of accurately retrieving short moments in practical scenarios.


In this study, we analyzed the challenges associated with existing methods in short-moment retrieval from both data and model perspectives. From a data perspective, we examined the feature distribution of short moments compared to other moments.  
As shown in \Cref{fig:method_feature_diversity}, the features of short moments tend to be more substantially concentrated than that of non‑short: the 95\% confidence ellipse of the covariance for short moments is markedly smaller. This observation shows that short moments exhibit relatively simple and less diverse feature distributions.
On the model side, we analyzed the trends in prediction accuracy by breaking down the model’s final output into center and length components, as illustrated in \Cref{fig:method_predictions}. Although conceptually, moments are defined by (start, end), existing models predict them in the format of (center, length). Interestingly, we found that the accuracy of both center and length predictions for short moments was significantly lower than for other moment types.


Based on the above analysis, we propose a new DETR-based MR framework that addresses the performance degradation when retrieving short moments. Our framework consists of two novel techniques:  a data augmentation technique called \textbf{MomentMix} and a Length-Aware Decoder (\textbf{LAD}). Through data analysis, we identified a key limitation in the feature diversity of short moments. It prompts us to design MomentMix, a two-stage mix-based augmentation strategy specifically to enhance the diversity of short-moment samples. In a video sample, we define the temporal moments relevant to the text query as the foreground and the unrelated moments as the background. MomentMix operates in two stages. In the first stage, we create new short foregrounds from rich foreground elements by cutting and mixing. In the next stage, to further diversify these short samples, we utilize portions of other videos as backgrounds, forming varied foreground-background combinations. This increased feature diversity of short moments leads to more robust detection of short moments.

Our analysis of model outputs revealed that both center and length prediction errors significantly contribute to performance drops in short-moment retrieval. To address this, we introduce a length-aware decoder designed to enhance center prediction by conditioning it on length. Specifically, we predefine length classes (e.g., short, middle, long) and uniformly assign each decoder query to these length-specific classes. Additionally, we modify the bipartite matching process so that queries are matched with ground-truth moments within the same length class. This approach improves the accuracy of both center and length predictions for short moments.



Our contributions are summarized as follows:
\begin{enumerate}
    \item We identify the root causes of performance degradation in short-moment retrieval for MR from both data and model perspectives.
    
    \item To address the issue of limited feature diversity in short moments, we propose a novel two-stage mix-based augmentation strategy, specifically tailored for short moment retrieval.
    
    \item To enhance center predictions for short moments, we introduced length conditioning into DETR-based MR methods for the first time, effectively creating  ``length-wise expert'' queries with length-wise matching.
    
    \item Our approach notably enhanced performance on the various MR datasets, resulting in significant improvement of mAP in \textsc{QVHighlights} (16.9\%; 39.84 $\rightarrow$ 46.61) and across other datasets.
    
\end{enumerate}

%% file: sections/02_related_work.tex

\section{Related Work}
\label{sec:related_work}

\subsection{Moment Retrieval}

Moment retrieval (MR) has been explored through DETR-based, non-DETR, and LLM-based approaches.
DETR-based methods~\cite{lei2021detecting_Moment-DETR} introduced a paradigm shift by formulating MR as a set prediction task, eliminating proposal dependency and extensive post-processing.
Recent works~\cite{moon2023query_QD-DETR, moon2023correlation_CG-DETR, liu2024towards_MESM, zhang2024temporally_TaskWeave, sun2024tr_TR-DETR, xiao2024bridging_UVCOM} further enhance video-text alignment.

Despite the advantages of DETR-based methods, recent non-DETR approaches~\cite{lin2023univtg_Univtg, liu2022umt_Umt, cao2024flashvtg} have introduced unified frameworks that integrate additional video temporal understanding tasks, including video summarization and highlight detection. 
While these methods achieve strong performance, they rely heavily on handcrafted heuristics such as Non-Maximum Suppression (NMS), making them computationally expensive and sensitive to hyperparameters.
LLM-based methods~\cite{ren2024timechat, huang2024vtimellm, jiang2024LLMEPET} leverage large-scale pre-trained models for improved textual understanding and reasoning.
However, their slow inference and high computational cost hinder their practicality in real-time applications.
Given the trade-offs across approaches, DETR-based models remain the practical solution for MR, balancing accuracy, efficiency, and scalability. 

Recent findings from FlashVTG~\cite{cao2024flashvtg} indicate performance degradation in short moments within the MR task. 
To the best of our knowledge, our study is the first to identify and analyze the underlying causes of this issue from both model and data perspectives, proposing novel approaches tailored for short moment performance improvement.


\begin{figure}[tb]
    \centering
    \includegraphics[width=0.90\linewidth]{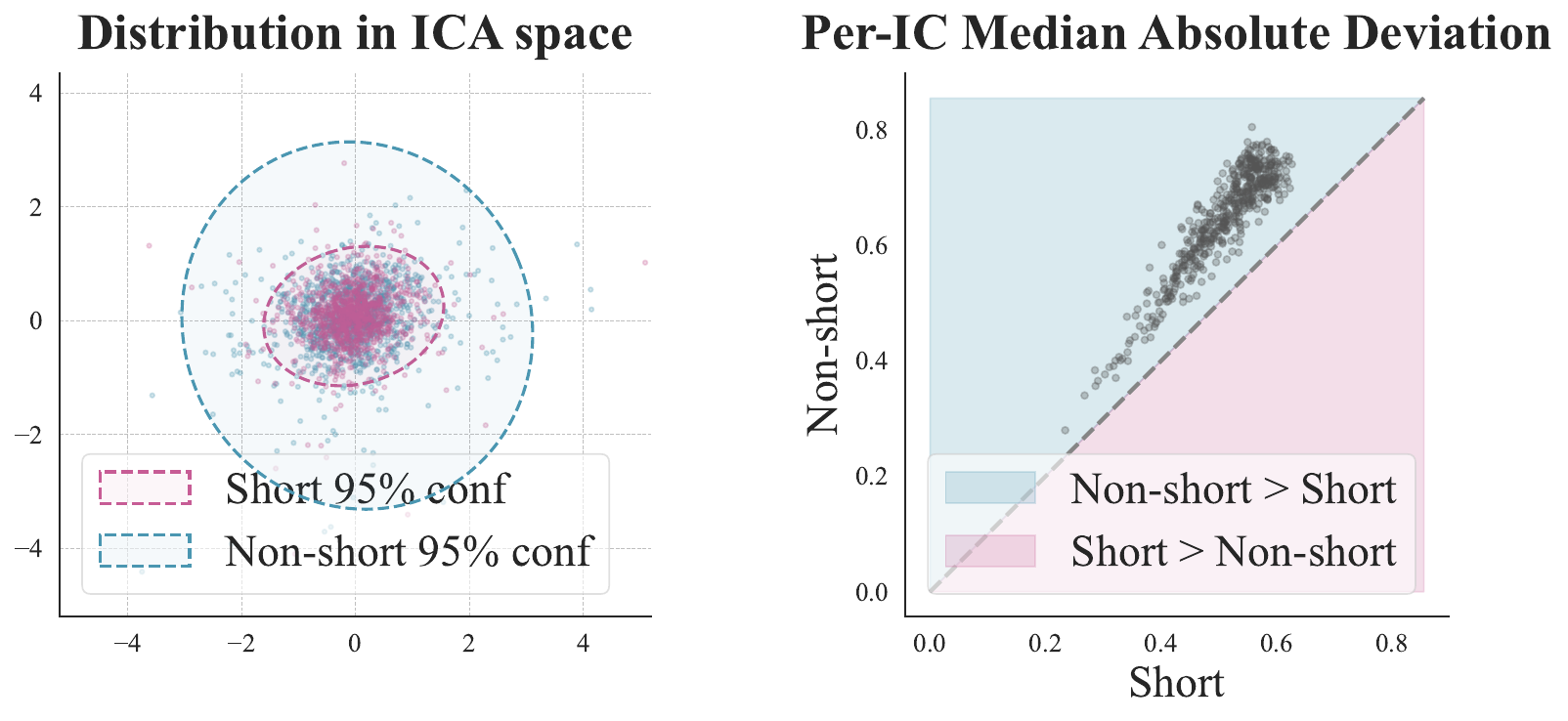}
    \vspace{-2mm}
    \caption{
        \textbf{Data Perspective Analysis}. We analyze \textsc{QVHighlights} \textit{val} features in ICA~\cite{hyvarinen2001ica}. [Left] short moments form a tighter cluster, as indicated by a smaller 95\% confidence ellipse of the covariance. [Right] For all Independent Components, the Median Absolute Deviation is greater for non-short moments, as shown by the majority of points lying above $y=x$.  Overall, these highlight short moments have the limited visual information.
    }
    \label{fig:method_feature_diversity}
\end{figure}

\subsection{Mixing-based Augmentation}
Mixing-based augmentations have been explored in both image and video tasks, each adapting spatial or temporal mixing based on task requirements.
In image classification, Mixup~\cite{zhang2017mixup} and CutMix~\cite{yun2019cutmix} create new image samples by interpolating or combining patches, promoting diverse feature representations. Copy-Paste~\cite{ghiasi2021copypaste} augments data for detection and segmentation by inserting objects from one image into another, increasing object and scene variety.
In video understanding tasks, VideoMix~\cite{yun2020videomix} inserts randomly selected video cuboid patches from one video into another, thereby introducing both spatial and temporal diversity. Similarly, VIPriors~\cite{kim2020VIPriors} extends traditional image-based mixing augmentations to the temporal dimension, which strengthens temporal feature representations and improves model robustness against temporal fluctuations.
However, these approaches primarily focus on modifying spatial features and are not directly applicable to the Moment Retrieval framework, which relies solely on frame-level features without spatial dimensions.
A related augmentation~\cite{liu2022reler} has been proposed for long-video moment retrieval, employing proposal-based methods by varying proposal lengths and inserting proposals into other videos. In contrast, our work explicitly targets the lack of feature diversity in short moments within a proposal-free DETR-based framework. To the best of our knowledge, we are the first to propose an augmentation strategy specifically tailored to short-moment retrieval in proposal-free methods.

%% file: sections/03_method.tex

\section{Method}
\label{sec:method}


\begin{figure}[tb]
    \centering
    \includegraphics[width=0.90\linewidth]{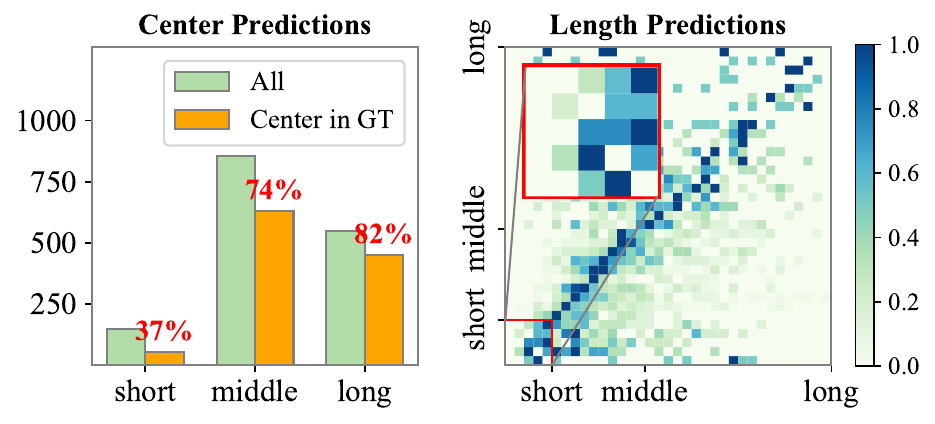}

    \caption{
        \textbf{Model Perspective Analysis}.     We analyze the top-1 predictions of QD-DETR on the \textsc{QVHighlights} \textit{val} set.     
        [Left] Percentage of predictions where the predicted center falls within the ground truth. Only 37\% of center predictions for short moments have their centers within the ground truth, indicating substantial errors in center prediction.
        [Right] Confusion matrix displaying predicted moment lengths (x-axis) versus ground-truth moment lengths (y-axis) across various durations. Short moments (highlighted in red) show a high rate of length prediction errors. 
    }
    \vspace{-3mm}
    \label{fig:method_predictions}
\end{figure}

\subsection{Motivation}


\noindent\textbf{Background.}
Suppose that a video consists of $\mathcal{N}_v$ clips, ${\{v_i\}}_{i=1}^{\mathcal{N}_v}$, and a text query of $\mathcal{N}_t$ words, ${\{t_i\}}_{i=1}^{\mathcal{N}_t}$.
The objective of moment retrieval (MR) is to predict a set of $\mathcal{N}_m$ moments, $\{m_i\}_{i=1}^{\mathcal{N}_m}$, corresponding to video clips relevant to the text query.
Each moment $m_i$ is defined by its center coordinate $c_i$ and length (span) $\sigma_i$, representing a contiguous subset of video clips. In this paper, we classify moments based on the following criteria: 1) A temporal moment within the video is defined as \textit{foreground} if it is relevant to the text query, and as \textit{background} if it is not. 2) Moments are categorized as short (less than 10 seconds), middle (10 to 30 seconds), or long (over 30 seconds) based on their temporal duration, consistent with the classification used in the previous method~\cite{lei2021detecting_Moment-DETR}.

\vspace{1mm}
\noindent\textbf{Why DETR-based Methods Underperform on Short Moments?} 
Recent approaches actively employ DETR~\cite{carion2020end_DETR} for the MR task because of its impressive performance and its end-to-end design that eliminates the need for extra post-processing.
We selected representative DETR-based models and analyzed their performance according to the length of target moments. Despite achieving strong performance, these models exhibited significant performance drops in retrieving short moments.
Specifically, for short moments, QD-DETR, TR-DETR, and UVCOM experienced mAP declines of 79.8\%, 78.0\%, and 72.4\%, respectively, compared to their overall performance. 
To address this degradation, we investigated the underlying causes from both data-centric and model-centric perspectives.


For the data-centric analysis, we examined the statistical characteristics of short moments. As illustrated in \textsc{QVHighlights}~\cite{lei2021detecting_Moment-DETR}, while the total number of short moments is comparable to that of other types of moments, the number of videos containing short moments is clearly limited. This led us to hypothesize that short moments might lack diverse contextual representation and could exhibit a narrow distribution in the data.  
To test this hypothesis, we compared the feature distribution of short moments to that of other moments. We analyze visual features in a fixed ICA~\cite{hyvarinen2001ica} coordinate system. As shown in \Cref{fig:method_feature_diversity}, on the selected independent components (ICs) the feature distribution of short moments was concentrated, indicating a significant lack of diversity. This pattern generalizes across other components, as all ICs have larger median absolute deviation for non-short moments. These observations suggest that short moments does not capture a wide range of visual features, leading to suboptimal generalization performance.

For the model-centric analysis, we evaluated the model's prediction tendencies by separately assessing the center and length predictions of short moments compared to other types of moments. As illustrated in \Cref{fig:method_predictions}, only 37\% of center predictions for short moments have their center within the ground truth, while 74\% for middle moments and 82\%  for long moments. This revealed that inaccuracies in center prediction are a significant source of overall error.

To overcome these limitations, we propose two novel techniques, MomentMix and Length-Aware Decoder that can be easily integrated into other DETR-based models. Our overall architecture follows the design of QD-DETR~\cite{moon2023query_QD-DETR}.



\subsection{MomentMix: Augmentation for Short Moment}

\label{subsec:MomentMix_Augmentation}

\begin{figure}[ht]
    \centering
    \includegraphics[width=1.00\linewidth]{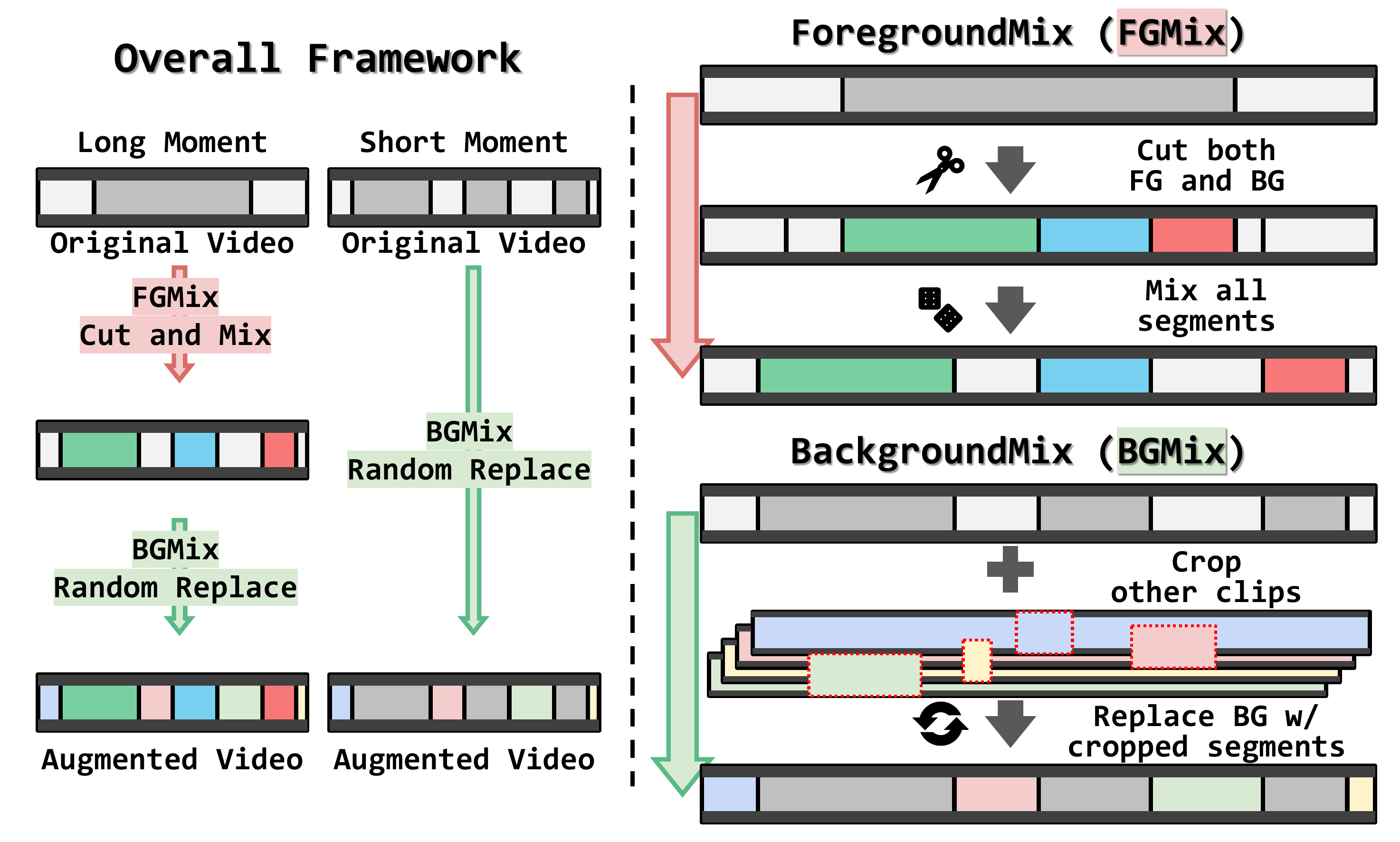}
    \caption{
        \textbf{MomentMix.}
        A two-stage mix-based data augmentation technique for generating short-moment samples.
        The first stage, \textit{ForegroundMix}, splits a long moment into shorter clips and shuffles them.
        The second stage, \textit{BackgroundMix}, preserves the foreground while replacing the background with randomly cropped temporal segments from other videos.
    }
    \label{fig:temp_aug}
\end{figure}

\noindent We propose MomentMix, a novel two-stage data augmentation strategy specifically designed to address the issue of low feature diversity for short moments. In the first stage (\textit{ForegroundMix}), diverse short foregrounds are generated by cutting and mixing longer foreground samples. In the second stage (\textit{BackgroundMix}), background diversity for augmented short samples is further enhanced by recombining backgrounds from various video clips. To the best of our knowledge, this is the first data augmentation approach specifically tailored for robust short moment retrieval.


\vspace{1mm}
\noindent\textbf{First Stage: ForegroundMix.} The goal of ForegroundMix is to increase the visual diversity of foreground features in short moments, enabling more generalized prediction. To achieve this, we randomly extract and mix rich foreground features from longer samples to create augmented short moments. Visual features within a single video naturally exhibit higher similarity compared to those from different videos. By exploiting features from other video clips, our method allows the model to generalize diverse contexts, enabling it to leverage a broader range of visual features for robust short-moment detection. 

Given an existing video training sample \(X = \{v_i\}_{i=0}^{\mathcal{N}_v}\) that contains a long foreground (moment) \( f_\text{source} = \{v_i\}_{i=s}^{e} \),  this foreground can be divided into sub-foregrounds \( f_1, f_2, \dots, f_n \) as follows:
\begin{equation}
f_\text{source} = \bigcup_{i=1}^{n} f_i, \:\: \text{where} \:\:  f_i \cap f_j = \emptyset \:\: \text{for all} \:\: i \neq j.
\end{equation}

Here, \( n = \frac{\texttt{len}(f_\text{source})}{{\varepsilon}_\text{cut}} \), where \(\varepsilon_\text{cut}\) is a hyperparameter determining the extent to which each sub-foreground is shortened relative to the original long foreground.

These sub-regions represent segments of the foreground, uniformly sampled as \( f_i^s, f_i^e \sim \text{Unif}(s, e) \), where $s$ and $e$ indicate the start and end of \(f_\text{source}\).
Similarly, the background region, $b_\text{source} = b_\text{front} \cup b_\text{back}$, is divided into \( n+1 \) sub-regions denoted as \( b_0, b_1, \dots, b_{n} \), representing the segments of the background as:
\begin{equation}
b_\text{source} = \bigcup_{i=0}^{n} b_i \:\: \text{where} \:\: b_i \cap b_j = \emptyset \:\: \text{for all} \:\: i \neq j.
\end{equation}
The original foreground, $\{f_i\}_{i=1}^{n}$, and backgrounds, $\{b_i\}_{i=0}^{n}$, are then shuffled as
\begin{equation}
\begin{gathered}
    \pi: \{f_1, f_2, \dots, f_n\} \rightarrow \{f_1', f_2', \dots, f_n'\}, \\
    \pi: \{b_0, b_1, \dots, b_n\} \rightarrow \{b_0', b_1', \dots, b_n'\}, \\
\end{gathered}
\end{equation} where $\pi$ be a random permutation function.
Each shuffled foregrounds $\{f'_i\}_{i=1}^{n}$ is then paired with backgrounds $\{b'_i\}_{i=0}^{n}$ to form the following augmented samples:
\begin{equation}
\begin{gathered}
    X' = b'_0 \cup \bigcup_{i=1}^{n} (f'_i \cup b'_{i}).
\end{gathered}
\end{equation}

To prevent temporal disruption during temporal mixing, we exclude queries involving explicit temporal relations from augmentation through preprocessing.

\vspace{1mm}


\noindent\textbf{Second Stage: BackgroundMix.} The goal of BackgroundMix is to improve the diversity of the visual background features of augmented short samples from ForegroundMix, thereby strengthening the association between foreground visual features and the text query. To achieve this, we keep the original foreground features while replacing the background with features from various videos' ground truth moment whose queries are semantically dissimilar to the current query. This method provides the model with richer training signals, allowing it to learn various boundaries more effectively for robust short moment retrieval.

A given the \(k\)-th video training sample \( X^k \), consists of \(\mathcal{N}^k_f\) foreground segments \( f^k = \{f^k_i\}_{i=1}^{\mathcal{N}^k_f} \) and \(\mathcal{N}^k_b\) background segments \( b^k = \{b^k_i\}_{i=1}^{\mathcal{N}^k_b} \). All segments within the video are defined as follows:
\begin{equation}
a^k =  f^k \cup b^k = \{a^k_i\}_{i=1}^{\mathcal{N}^k_a},
\text{where } \mathcal{N}^k_a = \mathcal{N}^k_f + \mathcal{N}^k_b.
\end{equation}

To increase feature diversity, we replace each background segment \(b^k_i\) of the \(k\)-th sample with a randomly cropped segment from a different training sample \(X^m\) (\(m \neq k\)). Specifically, for each \(b^k_i\), a segment \(a^m_j\) is randomly selected from \(X^m\) and cropped to match the duration of \(b^k_i\). The replacement is performed as follows:
\[
b^k_i \leftarrow \text{Crop}(a^m_j, \lvert b^k_i \rvert)
\]

To avoid introducing query-related content into the background, we replace backgrounds only with segments from other videos’ ground-truth moments whose queries are semantically dissimilar to the current query.




\subsection{Length-Aware Decoder}
\label{subsec:Length_Aware_Decoder}

\begin{figure}[ht]
    \centering
    \includegraphics[width=1.00\linewidth]{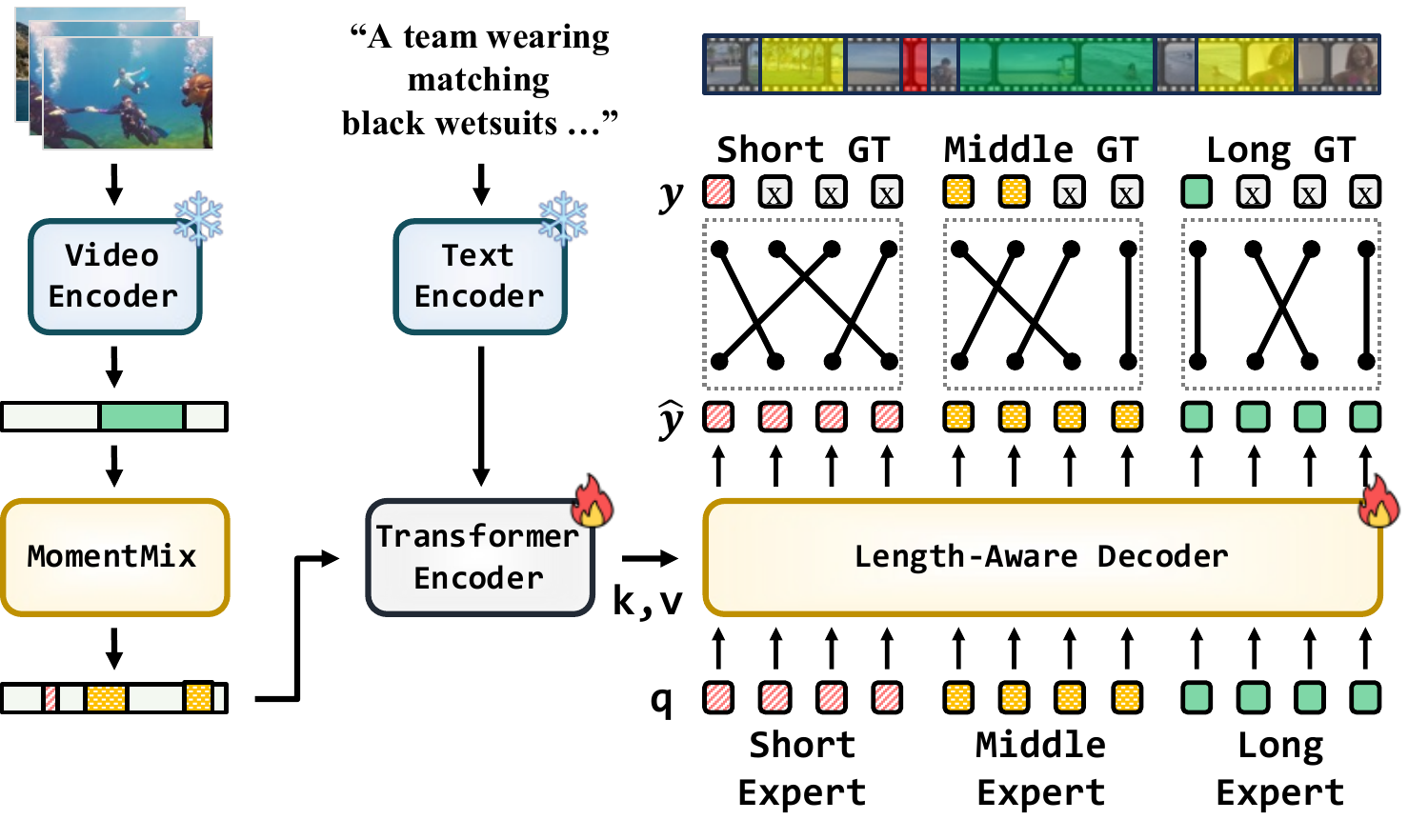} 
    \caption{
    \textbf{Overview of the Length-Aware Decoder (LAD).}
    The model applies a length-wise bipartite matching strategy, where predictions and ground truths of the same class are exclusively paired.
    This facilitates the creation of class-specific decoder queries, enhancing sequence modeling performance.
    }
    \label{fig:matching}
\end{figure}

\noindent 
In our previous analysis, we identified that the model struggles to accurately predict both the center and the length of short moments. To address this issue, we propose Length-Aware Decoder (LAD) that conditions the moment length, enabling the model to focus more effectively on center prediction. We categorize moment lengths into distinct classes—such as short, middle, and long—by analyzing a cumulative mAP graph and identifying inflection points as boundaries. (Detailed information can be found in the supplementary materials.) The decoder queries are trained using a length-wise matching approach based on these length categories. This categorization creates length-wise expert queries that better handle the specific characteristics of different moment lengths.

\vspace{1mm}
\noindent\textbf{Decoder Queries with Class-Pattern.} We define $\mathcal{N}_c$ as the number of length classes for assigning roles to decoder queries. Drawing inspiration from Anchor-DETR~\cite{wang2109anchor_Anchor_DETR}, we interpret \textit{pattern} in pattern embedding as a length category and create \textit{class-pattern embeddings} \(Q_c\) with dimension $d$:
\begin{equation}
    Q_c = \texttt{Embedding}(\mathcal{N}_c, d) \in \mathbb{R}^{\mathcal{N}_c \times d}.
\end{equation}
By replicating each class-pattern embedding $\mathcal{N}_q$ times (the number of queries per length), we obtain class-specific queries \(Q \in \mathbb{R}^{\mathcal{N}_c \mathcal{N}_q \times d}\). This approach ensures that decoder queries share the same class embedding within each length category, enabling each query to perform roles tailored to its specific length class.


\vspace{1mm}
\noindent\textbf{Length-wise Matching.}
To create the length-wise expertise within class-pattern embeddings, we revised the bipartite matching approach to operate on a per-class basis.
This method ensures that class-specific queries are matched and trained only with ground truth moments of the corresponding length class.
By categorizing ground truth moments into length classes and performing length-class-wise matching, we ensure precise alignment.
Although this may resemble group-wise matching in object detection~\cite{chen2023group_Group_DETR}, it differs significantly. As shown in \Cref{fig:matching}, existing methods use the same labels across all groups, resulting in one-to-many label assignments.
In contrast, our approach assigns a unique subset of labels to each length class, enabling one-to-one assignments and effectively creating a "length-wise expert" for matching.


We denote $\hat y = \{\hat y_i \}^{\mathcal{N}_c \mathcal{N}_q}_{i=1}$ as all the predicted moments from the decoder head, where $\mathcal{N}_c$ and $\mathcal{N}_q$ are a number of classes and a number of queries for each class $k \in \texttt{length-classes}$, respectively.
Then, the predictions belonging to class $k$ can be denoted as:
\begin{equation}
    \label{equation_ground_truth}
    \hat y^{(k)}  = \{\hat y_i \: \mid \: i^{th}\ query \in \texttt{class}\ k \}, \\
\end{equation}
When all ground truth moments are denoted as \(y = \{y_i\}^{\mathcal{N}_y}_{i=1}\), ground truth moments belonging to a specific class $k$ can be defined as:
\begin{equation}
    \label{equation_ground_truth2}
    y^{(k)} = \{ y_i \: \mid \:  \texttt{length}(y_i)  \: \in \: \texttt{class} \: k \: \}. \\
\end{equation}
For bipartite matching, by applying background \(\varnothing\) padding to make each set size $\mathcal{N}_q$, the final ground truth set becomes $\tilde y^{(k)} = \{{y_i^{(k)}}\}_{i=1}^{\mathcal{N}_q}$. 
The bipartite matching for each class $k$ is determined by finding the lowest cost among permutations of $\mathcal{N}_q$ elements, denoted as $\sigma \in {\mathfrak{S}}_{\mathcal{N}_q}$.
\begin{equation}
    \label{equation_ground_truth3}
    \hat \sigma^{(k)} = \argmin_{\sigma \in {\mathfrak{S}}_{\mathcal{N}_q}} \sum_{i}^{\mathcal{N}_q} \{ {{\mathbb{C}_{match}} (\tilde y^{(k)}_i, \hat y^{(k)}_{\sigma(i)})} \},
\end{equation}  where \({\mathbb{C}_{match}}\) is a \textit{matching cost} between ground truth and prediction. The matching cost function is set identically to that of the previous method~\cite{lei2021detecting_Moment-DETR}. 

This approach of class-wise matching aids in explicitly determining the length classes that were implicitly carried by the moment queries. By combining all the results from the bipartite matching for each class, we achieve an efficient matching that considers moment length.

%% file: sections/04_experiments.tex
\section{Experiments}
\label{sec:experiments}


\subsection{Experimental Setup}
\label{sec:experiment_setup}

\noindent\textbf{Datasets.}
We utilized three datasets, {\textsc{QVHighlights}} \cite{lei2021detecting_Moment-DETR}, \textsc{Charades-STA} \cite{gao2017tall_dataset_charades}, and \textsc{TACoS} \cite{regneri2013grounding_dataset_tacos}) for evaluation.
\textsc{QVHighlights} consists of over 10k YouTube videos covering various topics such as everyday activities, travel, social activities, and political activities.
It contains moments of various lengths distributed evenly and allows for testing our intended aspects effectively, as multiple moments appear within a single video.
Considering the diversity and complexity of the dataset, it covers the most realistic and challenging scenario. 
\textsc{Charades-STA} focuses on daily indoor activities, comprising 9,848 videos with 16,128 annotated queries.
The lengths of moments are mostly below 20 seconds.
\textsc{TACoS} primarily features activities in the cooking domain, consisting of 127 videos with 18,818 queries.
The video lengths vary from very short to nearly 800 seconds, with most moments being shorter than 30 seconds.

\vspace{1mm}
\noindent\textbf{Evaluation Metrics.}
Following the metrics of existing methods, we use mean average precision (mAP) with Intersection of Union (IoU) thresholds of 0.5 and 0.75, as well as the average mAP over multiple IoU thresholds [0.5: 0.05: 0.95].
Additionally, we report the standard metric Recall@1 (R1) metric, commonly used in single-moment retrieval, with IoU thresholds of 0.5 and 0.7. Also, we report the average R1 over multiple IoU thresholds [0.5: 0.05: 0.95].

\vspace{1mm}
\noindent\textbf{Implementation Details.}
We divided the classes based on the point where the change in performance was most significant in the validation results of each baseline model. To achieve this, we plotted the cumulative mAP graph with respect to length and identified the inflection points.
The thresholds for class division were then determined by calculating the k-means centers of these inflection points.
As a result, we set the thresholds for \textsc{QVHighlights} as [12, 36, 65, \(\inf\)], for \textsc{Charades-STA} as [5.67, 14, \(\inf\)], and for \textsc{TACoS} as [10, 19, 38, \(\inf\)], using UVCOM as baseline. In MomentMix, we set \(\varepsilon_\text{cut}=5\) for \textsc{QVHighlights} and \(\varepsilon_\text{cut}=10\) for \textsc{TACoS} and \textsc{Charades-STA}.  In LAD, the number of queries per class $ \mathcal{N}_q$ was set to 10.  For a fair comparison, we utilize the same features that previous works used. 
On \textsc{QVHighlights} and \textsc{TACoS}, the video features are extracted from SlowFast~\cite{feichtenhofer2019slowfast_slowfast} and CLIP visual encoder~\cite{radford2021learning_CLIP}.
On \textsc{Charades-STA}, we use two feature types as in previous works.
The first type is the video features from SlowFast and CLIP visual encoder, and text features extracted from the CLIP text encoder.
The second type is video features extracted from VGG \cite{simonyan2014very_vgg} and text features extracted from GloVe \cite{pennington2014glove_glove}. 
The model is trained for 200 epochs on all datasets with learning rates 1e-4. The batch size is 32 for \textsc{QVHighlights}, 8 for \textsc{Charades-STA}, and 16 for \textsc{TACoS}, following previous methods. We kept all the baseline parameters.


\subsection{Results}\label{sec:results}

We applied our method to QD-DETR~\cite{moon2023query_QD-DETR}, a common baseline in many studies.
However, since our method can be easily added to other models, we further validated our method on three recent methods (TR-DETR~\cite{sun2024tr_TR-DETR}, and UVCOM~\cite{xiao2024bridging_UVCOM}) to demonstrate its effectiveness.
We compared our approach against existing moment retrieval methods, including the latest DETR-based models.
While existing models report only overall performance, we also analyze the performance of each length.

\vspace{2mm}

\noindent\textbf{Performance with respect to Moment Length on \textsc{QVHighlights}.} 
In \Cref{tab:exp_length_test}, our method significantly improves short-moment performance across all baselines. Specifically, for QD-DETR, the R1 average and mAP average for short moments increased by +6.38\%p and +6.81\%p, respectively. Moreover, our approach consistently outperforms all baselines in mAP average across all lengths.


\input{tables/method_qv_length_test}

\noindent\textbf{Overall Performance on \textsc{QVHighlights}.} 
In \Cref{tab:exp_qv_test}, our method yields significant improvements across all metrics, indicating enhanced overall performance across all baselines. Notably, while our primary objective was to improve the short moment performance in Moment Retrieval (MR) by enhancing feature diversity, we also observed substantial performance gains in Highlight Detection (HD). This demonstrates that enhancing feature diversity is an effective strategy that can positively impact other tasks.


\input{tables/method_qv_test}

\vspace{4mm}
\noindent\textbf{Overall Performance on \textsc{Charades-STA} and \textsc{TACoS}.} 
As shown in~\Cref{tab:exp_cha_tacos}, our method significantly improves performance on \textsc{Charades-STA}, increasing R1@0.7 by +3.58\%p with SlowFast and CLIP features, and by +1.97\%p with VGG features. Furthermore, our method achieves notable gains (+5.82\%p in R1@0.5) on \textsc{TACoS}, which encompasses moments with a broader range of lengths compared to \textsc{QVHighlights}. This substantial improvement demonstrates the strong generalization capability of our length-aware approach.


\input{tables/method_other_dataset}

\begin{figure}[h]
    \centering
    \includegraphics[width=\linewidth]{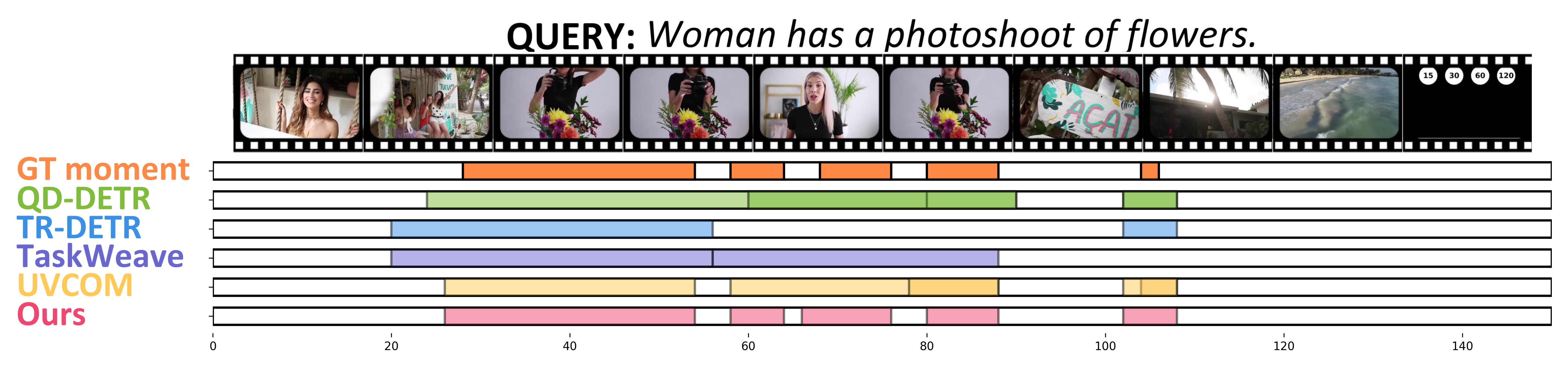}
    \includegraphics[width=\linewidth]{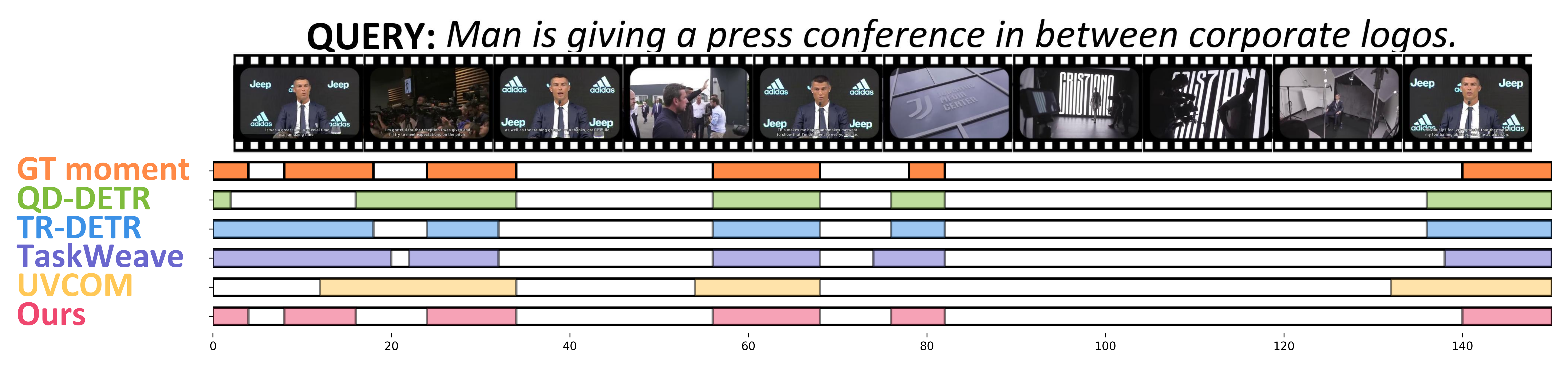}
    \caption{
        {Qualitative results on \textsc{QVHighlights}. Existing models often fail to accurately distinguish between foreground and background, leading to unsuccessful predictions or missed detections of short moments. In contrast, our model is capable of predicting short moments with greater accuracy and robustness. }
    }
    \vspace{-2mm}
    \label{fig:exp_qualitative}
\end{figure}

\vspace{-2mm}
\noindent\textbf{Qualitative Results.} 
We visualized predictions with confidence scores exceeding 0.7, using an alpha value of 0.5. As shown in \Cref{fig:exp_qualitative}, by applying our method, short moments predicted as background in other methods can now be accurately captured. Also, predictions that merged multiple short instances into a single long instance can now be segmented into precise, fine-grained predictions.

\subsection{Ablation Studies and Discussions}\label{sec:Ablation studies}

\noindent\textbf{Component Analysis.} In \Cref{tab:exp_component_analysis}, we examined the impact of MomentMix and the Length-Aware Decoder on enhancing performance for short moments, observing overall gains.
While each component individually improves performance, their combined application leads to even greater improvements.
This suggests that our two components, MomentMix and LAD, each contribute effectively without redundancy, making their combined use the effective approach for tackling the challenge of short-moment retrieval.


\input{tables/method_component_analysis}


\noindent\textbf{Effect of MomentMix.}
To validate the effectiveness of MomentMix, a novel mixing-based augmentation for moment retrieval (MR) for short moment, we compare it with four baseline augmentations: \textit{Gaussian Noise}, \textit{Random Drop},  \textit{Random Crop} and \textit{Random Insert}.
For Gaussian Noise and Random Drop, we report only the best-performing hyperparameters: adding Gaussian noise with a standard deviation of 0.01 and masking 50\% of frame features, respectively.
Random Crop and Random Insert were originally proposed for proposal-based long-video MR and adapted by us for proposal-free DETR-based methods. In contrast, our proposed MomentMix uniquely employs foreground mixing (\textit{ForegroundMix}) and background replacement (\textit{BackgroundMix}). Specifically, ForegroundMix cuts and mixes foreground segments rather than simply cropping them, while BackgroundMix replaces each background segment independently with clips from different videos, instead of inserting a single foreground segment.
As shown in~\Cref{tab:supple_mmix}, MomentMix consistently outperforms baseline augmentations, significantly boosting performance on short moments, while also improving longer moments.

\vspace{-1mm}

\input{tables/supple_eff_momentmix}

\vspace{-1mm}

\noindent\textbf{Evaluation in Few-shot Scenarios.} To validate the effectiveness of MomentMix as a data augmentation technique, we conducted experiments using 50\%, 20\%, and 10\% of the training data.
As shown in \Cref{tab:exp_semi}, our method outperformed the baseline (QD-DETR) with substantial performance gains. Specifically, utilizing only half of the training samples with our method surpassed the baseline performance that used the entire dataset. Additionally, even in the extreme scenario of using just 10\% of the training samples, our method achieved remarkable improvements of +9.71\%p in the R1 average and +8.45\%p in the mAP average. These results indicate that MomentMix effectively generates new training samples by enhancing feature diversity.

\vspace{-1mm}

\input{tables/method_semi}

\vspace{-1mm}

\begin{figure}[h]
    \centering
    \includegraphics[width=\linewidth]{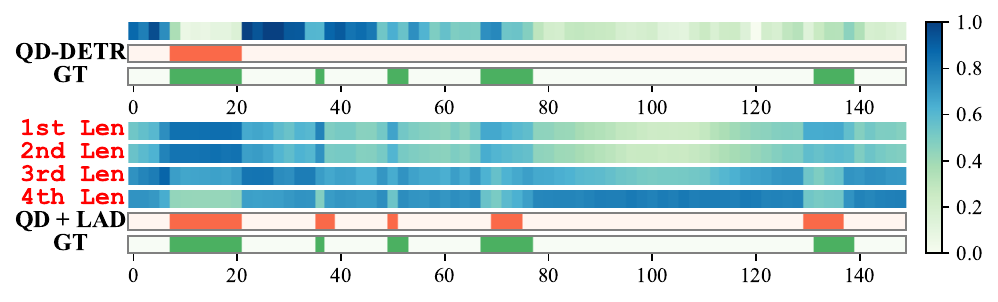}

    \vspace{-2mm}
    \caption{
        {Decoder query attention to encoder features  on \textsc{QVHighlights}. The top figure shows baseline attention, while the bottom shows attention after applying LAD. Length-expert queries are highlighted in red text. LAD effectively aligns attention patterns with the corresponding length characteristics.}
    }
    \label{fig:lad_attn}
\end{figure}

\noindent\textbf{Attention in LAD.}
To examine whether our LAD attends to the characteristics of each length-specific query as intended, we investigated attention between these queries and encoder features. In \cref{fig:lad_attn}, the existing method \cite{moon2023query_QD-DETR} predicts moments by focusing on the boundary regardless of length, which we refer to as \textit{boundary attention} of the moment. However, when applying our method, we observed that attention varies according to the characteristics of each length-specific query. For the first length class (shorter moments), attention is focused more on the inside of the moment rather than the boundary, which we refer to as  \textit{inside attention} of the moment.  On the other side, for the fourth length class (longer moments), attention is directed toward the boundary similar to the existing method. This indicates that our query attends to the length rather than the center of the moment appears to be more efficient for detection, hence the boundary attention. Conversely, for short moments, focusing on the center rather than the length is more efficient, resulting in a moment of inside attention.

%% file: tables/method_qv_length_test.tex
\begin{table}[ht]

\setlength{\tabcolsep}{0.3em}
\centering

\vspace{-1mm}

\begin{small}
\setlength{\tabcolsep}{2pt}

\resizebox{\linewidth}{!}
{

    \begin{tabular}{
        l    c c   c c   c c    c c
    }

    \toprule

    \multirow{3}{*}{Method} & \multicolumn{2}{c}{\textbf{Short}} & \multicolumn{2}{c}{Middle} & \multicolumn{2}{c}{Long}  & \multicolumn{2}{c}{Full} \\

    \cmidrule(l){2-3}  \cmidrule(l){4-5}  \cmidrule(l){6-7}  \cmidrule(l){8-9}

      & R1 & mAP & R1 & mAP & R1 & mAP  & R1 & mAP \\

    

    \midrule

    CG-DETR~\cite{moon2023correlation_CG-DETR}  & 5.70	 & 9.61	 & 42.25	 & 45.23	 & 44.58	 & 64.35	 & 44.10	 & 42.86	\\
    TaskWeave$\ddagger$~\cite{zhang2024temporally_TaskWeave} & 2.91	 & 6.31	 & 38.93	 & 44.00	 & 47.58	 & 52.67	 & 42.45	 & 43.33	\\

    $R^2$-Tuning$\ddagger$~\cite{liu2024r2-tuning} & 6.83	 & 11.96	 & 38.86	 & 46.00	 & 49.56	 & 54.23	 & 44.16	 & 46.10	\\
    FlashVTG~\cite{cao2024flashvtg} & \underline{10.35}	 & \underline{14.84}	 & 41.62	 & \underline{49.04}	 & 47.09	 & 51.68	 & 45.84	 & 47.59	\\


    \hline

    QD-DETR~\cite{moon2023query_QD-DETR}	 & 3.95	 & 6.98	 & 37.39	 & 41.12	 & 42.86	 & 46.95	 & 40.01	 & 39.84	\\
    \rowcolor{olive!10} ~ +Ours	 & {10.33}	 & 13.79	 & 41.32	 & 46.94	 & 45.40	 & 52.45	 & 45.02	 & 46.61	\\
	\vspace{-4pt} & \gainp{+6.38} 	& \gainp{+6.81} 	& \gainp{+3.93} 	& \gainp{+5.82} 	& \gainp{+2.54} 	& \gainp{+5.50} 	& \gainp{+5.01} 	& \gainp{+6.77} 	\\

    \hline
    
    TR-DETR~\cite{sun2024tr_TR-DETR}	 & 4.95	 & 8.22	 & 40.08	 & 43.27	 & \textbf{47.63}	 & 50.80	 & 43.70	 & 42.62	\\
    \rowcolor{olive!10} ~ +Ours	 & 10.30	 & {14.57}	 & \underline{42.54}	 & {47.90}	 & \underline{47.61}	 & \textbf{53.58}	 & \underline{46.59}	 & \underline{47.77}	\\
	\vspace{-4pt} & \gainp{+5.35} 	& \gainp{+6.35} 	& \gainp{+2.45} 	& \gainp{+4.63} 	& \gainm{-0.02} 	& \gainp{+2.78} 	& \gainp{+2.89} 	& \gainp{+5.15} 	\\
    \hline

    UVCOM~\cite{xiao2024bridging_UVCOM}	 & 5.28	 & 10.67	 & 41.81	 & 44.90	 & 44.95	 & 48.37	 & 43.85	 & 43.18	\\
    \rowcolor{olive!10} ~ +Ours	 & \textbf{11.58}	 & \textbf{16.35}	 & \textbf{43.22}	 & \textbf{52.62}	 & 46.48	 & \underline{52.62}	 & \textbf{46.92}	 & \textbf{48.21}	\\
	\vspace{-4pt} & \gainp{+6.31} 	& \gainp{+5.68} 	& \gainp{+1.41} 	& \gainp{+7.72} 	& \gainp{+1.53} 	& \gainp{+4.25} 	& \gainp{+3.07} 	& \gainp{+5.03} 	\\

    \bottomrule
    \end{tabular}

}

\end{small}

\vspace{-1mm}
\caption{Performance gains of our method on the \textsc{QVHighlights} \textit{test} set across different moment lengths. {\ddag} means test results from checkpoint provided by authors.}

\label{tab:exp_length_test}

\end{table}

%% file: tables/method_qv_test.tex
\begin{table}[t]

\setlength{\tabcolsep}{0.3em}
\centering

\vspace{-1mm}
\begin{small}
\setlength{\tabcolsep}{2pt}

\resizebox{\linewidth}{!}
{

    \begin{tabular}{
         l  c c c  c c c  c c
    }

    \toprule

\multirow{3}{*}{Method} & \multicolumn{6}{c}{\textbf{MR}}  & \multicolumn{2}{c}{\textbf{HD}} \\

\cmidrule(l){2-7} \cmidrule(l){8-9}

& \multicolumn{3}{c}{R$1$}  & \multicolumn{3}{c}{mAP} & \multicolumn{2}{c}{$\geq$ Very Good} \\

\cmidrule(l){2-4} \cmidrule(l){5-7} \cmidrule(l){8-9}

 & @$0.5$ & @$0.7$ & Avg. & @$0.5$ & @$0.75$ & Avg. & mAP & HIT@$1$ \\

    \midrule

M-DETR~\cite{lei2021detecting_Moment-DETR}	 & 52.89	 & 33.02	& -	 & 54.82	 & 29.40	 & 30.73	 & 35.69	 & 55.60	\\
UMT $\dagger$~\cite{liu2022umt_Umt}	 & 56.23	 & 41.18	& -	 & 53.83	 & 37.01	 & 36.12	 & 38.18	 & 59.99	\\
EaTR~\cite{jang2023knowing_EaTR}	 & 57.98	 & 42.41	& -	 & 59.95	 & 39.29	 & 39.00	& -	& -	\\
UniVTG~\cite{lin2023univtg_Univtg}	 & 58.86	 & 40.86	& -	 & 57.60	 & 35.59	 & 35.47	 & 38.20	 & 60.96	\\
CG-DETR~\cite{chen2023group_Group_DETR}  & 65.43	 & 48.38	 & 44.10	 & 64.51	 & 42.77	 & 42.86	 & 40.33	 & \underline{66.21} \\
MomentDiff~\cite{li2024momentdiff_momentdiff}	 & 57.42	 & 39.66	& -	 & 54.02	 & 35.73	 & 35.95	& -	& -	\\
TaskWeave$\ddagger$~\cite{zhang2024temporally_TaskWeave}	 & 61.87	 & 46.24	 & 42.45	 & 63.75	 & 43.63	 & 43.33	 & 37.87	 & 59.08	\\
BAM-DETR~\cite{lee2025bam-detr}	 & 62.71	 & 48.64	& -	 & 64.57	 & 46.33	 & 45.36	& -	& -	\\


$R^2$-Tuning$\ddagger$~\cite{liu2024r2-tuning} & \textbf{66.08}	 & 48.90	 & 44.16	 & \textbf{68.09}	 & 47.65	 & 46.10	 & 39.18	 & 64.20 \\

FlashVTG~\cite{cao2024flashvtg} &	 \textbf{66.08} & 50.00 & 45.84 & \underline{67.99} & 48.70 & 47.59 & \underline{41.07} & {66.15} \\

\hline
QD-DETR~\cite{moon2023query_QD-DETR}	 & 61.22	 & 44.49	 & 40.01	 & 62.31	 & 39.45	 & 39.84	 & 39.01	 & 62.13	\\
\rowcolor{olive!10} ~ +Ours	 & 63.62	 & 48.77	 & 45.02	 & 65.50	 & 47.78	 & 46.61	 & 40.35	 & 64.46	\\
	\vspace{-4pt} & \gainp{+2.40} 	& \gainp{+4.28} 	& \gainp{+5.01} 	& \gainp{+3.19} 	& \gainp{+8.33} 	& \gainp{+6.77} 	& \gainp{+1.34} 	& \gainp{+2.33} 	\\

\hline
    
TR-DETR~\cite{sun2024tr_TR-DETR}	 & {64.66}	 & 48.96	 & 43.70	 & 63.98	 & 43.73	 & 42.62	 & 39.91	 & 63.42	\\
\rowcolor{olive!10} ~ +Ours	 & \underline{65.63}	 & \textbf{51.23}	 & \underline{46.59}	 & 66.89	 & \underline{49.04}	 & \underline{47.77}	 & \textbf{41.54}	 & 66.02	\\
	\vspace{-4pt} & \gainp{+0.97} 	& \gainp{+2.27} 	& \gainp{+2.89} 	& \gainp{+2.91} 	& \gainp{+5.31} 	& \gainp{+5.15} 	& \gainp{+1.63} 	& \gainp{+2.60} 	\\

\hline
    
UVCOM~\cite{xiao2024bridging_UVCOM}	 & 63.55	 & 47.47	 & 43.85	 & 63.37	 & 42.67	 & 43.18	 & 39.74	 & 64.20	\\
\rowcolor{olive!10} ~ +Ours	 & 65.37	 & \underline{50.71}	 & \textbf{46.92}	 & 66.65	 & \textbf{49.22}	 & \textbf{48.21}	 & 40.91	 & \textbf{66.54}	\\
	\vspace{-4pt} & \gainp{+1.82} 	& \gainp{+3.24} 	& \gainp{+3.07} 	& \gainp{+3.28} 	& \gainp{+6.55} 	& \gainp{+5.03} 	& \gainp{+1.17} 	& \gainp{+2.34} 	\\

    \bottomrule
    \end{tabular}
}

\end{small}

\vspace{-1mm}

\caption{Performance comparison on \textsc{QVHighlights} \textit{test} set. {\dag} indicates training with additional audio features. {\ddag} means test results from checkpoint provided by authors.
}

\vspace{-3mm}

\label{tab:exp_qv_test}

\end{table}

%% file: tables/method_other_dataset.tex
\begin{table}[ht]

\centering

\vspace{-1mm}

\setlength{\tabcolsep}{0.05pt}

\begin{small}

\resizebox{0.9\linewidth}{!}
{

    \begin{tabular}{
        l c   c c   c c   c c
    }
    \toprule
    \multirow{2}{*}{Method} & \multicolumn{2}{c}{\textsc{Charades-STA}} & \multicolumn{2}{c}{\textsc{TACoS}} & \multicolumn{2}{c}{\textsc{Charades-STA}$^{\ddag}$}\\
    \cmidrule(l){2-3}  \cmidrule(l){4-5}  \cmidrule(l){6-7}
    & R1@0.5 & R1@0.7 & R1@0.5 & R1@0.7 & R1@0.5 & R1@0.7\\

    \midrule
SAP~\cite{gao2021fast_SAP}          & -              & -              & -          & -          & 27.42          & 13.36 \\
SM-RL~\cite{wang2019language_sm_rl}        & -              & -              & -          & -          & 24.36          & 11.17 \\
MAN~\cite{Zhang_2019_CVPR_man}          & -              & -              & -          & -          & 41.24          & 20.54 \\
2D-TAN~\cite{zhang2020learning_2d_tan}       & 46.02          & 27.50          & 27.99      & 12.92      & 40.94          & 22.85 \\
VSLNet~\cite{zhang2020span_vslnet}       & 42.69          & 24.14          & 23.54      & 13.15      & -              & - \\
M-DETR~\cite{lei2021detecting_Moment-DETR}       & 53.63          & 31.37          & 24.67      & 11.97      & -              & - \\
QD-DETR~\cite{moon2023query_QD-DETR}      & 57.31          & 32.55          & -          & -          & 52.77          & 31.13 \\
UniVTG~\cite{lin2023univtg_Univtg}       & 58.01          & 35.65          & 34.97      & 17.35      &                &  \\
TR-DETR~\cite{sun2024tr_TR-DETR}      & 57.61          & 33.52          & -          & -          & 53.47          & 30.81 \\


BAM-DETR~\cite{lee2025bam-detr}        & 59.83          & \underline{39.83}          & {41.54}      & \underline{26.77}      & -  & - \\

$R^2$-Tuning~\cite{liu2024r2-tuning} & - & - & 38.72 & 25.12 & - & - \\

FlashVTG~\cite{cao2024flashvtg} & \underline{60.11} & 38.01 & \underline{41.76} & 24.74 & {54.25} & \textbf{37.42} \\

\hline

UVCOM~\cite{xiao2024bridging_UVCOM}        & 59.25          & 36.64          & 36.39      & 23.32      & \underline{54.57}          & {34.13} \\

\rowcolor{olive!10}~+Ours & \textbf{61.45}    & \textbf{40.22}   & \textbf{42.21} & \textbf{28.02}   & \textbf{56.16} & \underline{36.10} \\

\vspace{-4pt} & \gainp{+2.20} 	& \gainp{+3.58} 	& \gainp{+5.82} 	& \gainp{+4.70} 	& \gainp{+1.59} 	& \gainp{+1.97} \\

    \bottomrule
    \end{tabular}
}

\end{small}

\vspace{-1mm}

\caption{Results on \textsc{Charades-STA} and \textsc{TaCoS} \textit{test} set. {\ddag} indicates training with VGG features and GloVe features.}

\label{tab:exp_cha_tacos}

\end{table}

%% file: tables/method_component_analysis.tex
\begin{table}[ht]

\setlength{\tabcolsep}{0.3em}
\centering

\vspace{-1mm}

\begin{small}
\setlength{\tabcolsep}{3.5pt}

\resizebox{\linewidth}{!}
{

    \begin{tabular}{
        c c    c c   c c   c c    c c
    }

    \toprule

    \multicolumn{2}{c}{} & \multicolumn{2}{c}{\textbf{Short}} & \multicolumn{2}{c}{Middle} & \multicolumn{2}{c}{Long}  & \multicolumn{2}{c}{Full} \\

    \cmidrule(l){3-4}  \cmidrule(l){5-6}  \cmidrule(l){7-8}  \cmidrule(l){9-10}

     \textit{MMix} & \textit{LAD} & R1 & mAP & R1 & mAP & R1 & mAP  & R1 & mAP \\

    \midrule

\nomark & \nomark	 & 4.57	 & 7.77	 & 38.89	 & 43.10	 & 42.62	 & 47.44	 & 41.06	 & 41.00	\\
\yesmark   & \nomark	 & 6.48	 & 11.44	 & 42.96	 & 46.88	 & 44.15	 & 49.57	 & 44.66	 & 44.68	\\
 \nomark & \yesmark	 & 8.76	 & 11.01	 & 40.55	 & 45.53	 & 43.69	 & 50.76	 & 43.65	 & 44.48	\\
\yesmark   & \yesmark 	 & \textbf{11.07}	 & \textbf{15.27}	 & \textbf{43.12}	 & \textbf{48.53}	 & \textbf{44.39}	 & \textbf{52.65}	 & \textbf{46.13}	 & \textbf{47.70}	\\

    \bottomrule
    \end{tabular}
}

\end{small}

\vspace{-2mm}

\caption{Performance comparison with baseline(QD-DETR) on \textsc{QVHighlights} \textit{val} set. MMix, and LAD indicate \underline{M}oment\underline{Mix} and \underline{L}ength-\underline{A}ware \underline{D}ecoder, respectively. 
}

\label{tab:exp_component_analysis}

\end{table}

%% file: tables/supple_eff_momentmix.tex
\begin{table}[ht]

\setlength{\tabcolsep}{0.3em}
\centering

\begin{small}
\setlength{\tabcolsep}{2pt}

\resizebox{\linewidth}{!}
{

    \begin{tabular}{
        l    c c   c c   c c    c c
    }

    \toprule

    \multirow{3}{*}{Method} & \multicolumn{2}{c}{\textbf{Short}} & \multicolumn{2}{c}{Middle} & \multicolumn{2}{c}{Long}  & \multicolumn{2}{c}{Full} \\

    \cmidrule(l){2-3}  \cmidrule(l){4-5}  \cmidrule(l){6-7}  \cmidrule(l){8-9}

      & R1 & mAP & R1 & mAP & R1 & mAP  & R1 & mAP \\    

    \midrule

Baseline	 & 4.57	 & 7.77	 & 38.89	 & 43.10	 & 42.62	 & 47.44	 & 41.06	 & 41.00	\\

\hline

Gaussian Noise	 & 5.20	 & 7.62	 & 37.97	 & 42.64	 & 43.54	 & 48.18	 & 41.01	 & 40.58	\\

Random Drop	 & 4.15	 & 7.24	 & 39.87	 & 44.05	 & 42.44	 & 47.24	 & 41.48	 & 40.95	\\

Random Crop \cite{liu2022reler}	 & 5.31	 & 8.99	 & 39.34	 & 43.03	 & 41.01	 & 46.00	 & 40.95	 & 41.00	\\

Random Insert \cite{liu2022reler}	 & \textbf{6.71}	 & 8.72	 & 38.59	 & 42.77	 & 41.69	 & 45.94	 & 41.12	 & 40.46	\\

\hline

\rowcolor{olive!10}MomentMix	 & 6.48	 & \textbf{11.44}	 & \textbf{42.96}	 & \textbf{46.88}	 & \textbf{44.15}	 & \textbf{49.57}	 & \textbf{44.66}	 & \textbf{44.68}	\\

    \bottomrule
    \end{tabular}
}

\end{small}

\vspace{-1mm}

\caption{
    Performance comparison of MomentMix and other augmentations—Gaussian Noise, Random Drop, Random Crop, and Random Insert—on the \textsc{QVHighlights} \textit{val} set with QD-DETR as the baseline.
    Other augmentations fail to improve performance, while MomentMix achieves significant gains in short moments and overall robustness in moment retrieval tasks.
}

\label{tab:supple_mmix}

\end{table}

%% file: tables/method_semi.tex
\begin{table}[htb]

    \centering

\vspace{-1mm}
\setlength{\tabcolsep}{4pt}
\resizebox{1.0\linewidth}{!}
{
    \begin{tabular}{
        l   c c c   c c c 
    }
    \toprule
    {\multirow{3}{*}{Method}} & \multicolumn{3}{c}{R1} & \multicolumn{3}{c}{mAP}  \\

    \cmidrule(l){2-4}  \cmidrule(l){5-7}  
    
     & @0.5 & @0.7 & Avg.  & @0.5 & @0.75 & Avg.  \\
    \midrule

100\% train data & 61.39	 & 46.18	 & 41.06	 & 61.68	 & 41.57	 & 41.00	\\
\hline
 50\% train data & 57.23	 & 40.26	 & 36.10	 & 57.51	 & 35.63	 & 35.98	\\
\rowcolor{olive!10} ~ + MomentMix	 & 63.16	 & 47.74	 & 43.36	 & 61.91	 & 41.90	 & 41.73	\\
	\vspace{-4pt} & \gainp{+5.93} 	& \gainp{+7.48} 	& \gainp{+7.26} 	& \gainp{+4.40} 	& \gainp{+6.27} 	& \gainp{+5.75} 	\\
\hline
    
 20\% train data& 46.84	 & 30.45	 & 26.58	 & 48.27	 & 25.35	 & 26.88	\\
\rowcolor{olive!10} ~ + MomentMix	 & 52.45	 & 37.68	 & 33.69	 & 52.66	 & 34.25	 & 33.72	\\
	\vspace{-4pt} & \gainp{+5.61} 	& \gainp{+7.23} 	& \gainp{+7.11} 	& \gainp{+4.39} 	& \gainp{+8.90} 	& \gainp{+6.84} 	\\
\hline
    
 10\% train data& 32.45	 & 16.84	 & 15.90	 & 37.10	 & 15.37	 & 18.17	\\
\rowcolor{olive!10} ~ + MomentMix	 & 43.10	 & 28.71	 & 25.61	 & 44.97	 & 26.12	 & 26.62	\\
	\vspace{-4pt} & \gainp{+10.65} 	& \gainp{+11.87} 	& \gainp{+9.71} 	& \gainp{+7.87} 	& \gainp{+10.75} 	& \gainp{+8.45} 	\\

        \bottomrule
    \end{tabular}
}

\vspace{-1mm}

\caption{
    Results on the \textsc{QVHighlights} \textit{val} set using 50\%, 20\%, and 10\% of the original training data.
}
\label{tab:exp_semi}

\end{table}

%% file: sections/05_conclusion.tex
\section{Discussion}
\noindent\textbf{Conclusion.}  
This study addressed the limitations of short-moment retrieval in existing DETR-based approaches from both data and model perspectives.
To mitigate the issue of limited feature diversity in short moments, we introduced MomentMix, a novel two-stage mix-based data augmentation.
These strategies enhance the feature representations of both foreground and background elements for short-moment data samples.
On the model side, we identified inaccuracies in center predictions for short moments and proposed a Length-Aware Decoder with a novel bipartite matching process conditioned on moment length.
This approach leverages length expert queries to improve center prediction accuracy.
Extensive experiments demonstrate that our method outperforms state-of-the-art DETR-based moment retrieval models in terms of R1 and mAP on benchmark datasets.
Moreover, our methodology can be seamlessly integrated with other DETR-based models, paving the way for future advancements in the field.

\vspace{2mm}
\noindent\textbf{Future Work.}  
In future work, we plan to explore adapting and extending our augmentation approach to non-DETR architectures. 
Additionally, to better handle datasets with strong temporal dependencies, we will explore more sophisticated temporal-aware augmentations.

\section*{Acknowledgments}

This research was supported by the Basic Science Research Program through the National Research Foundation of Korea (NRF) funded by the MSIP (No. RS-2025-00520207 and No. RS-2023-00240135); the Institute of Information \& communications Technology Planning \& Evaluation (IITP) grant funded by the Korea government (MSIT) (No. RS-2024-00457882, National AI Research Lab Project and RS-2019-II190075, Artificial Intelligence Graduate School Program(KAIST)); and the IITP grant funded by MSIT and KEIT grant funded by the Korea government (MOTIE) (No. 2022-0-01045, No. 2022-0-00680, and No. RS-2025-02217259, Development of self-evolving AI bias detection-correction-explain platform based on international multidisciplinary governance, 14.3\%).
We thank Hyogon Ryu and Jihwan Eom for their valuable discussions.

%% file: sections/06_supplementary_materials.tex


    




\clearpage
\newpage
\setcounter{page}{1}
\maketitlesupplementary
\appendix

\startcontents[supplement] 
\printcontents[supplement]{}{1}{\section*{\contentsname}}  

\setcounter{section}{0}

\setcounter{figure}{0}
\renewcommand{\thetable}{A\arabic{table}}

\setcounter{table}{0}
\renewcommand{\thefigure}{A\arabic{figure}}

\setcounter{equation}{0}
\renewcommand{\theequation}{A\arabic{equation}}



\section{Additional Ablation Studies}
\label{sec:supple_ablation}


\begin{figure}[b]
    \centering
    \includegraphics[width=0.93\linewidth]{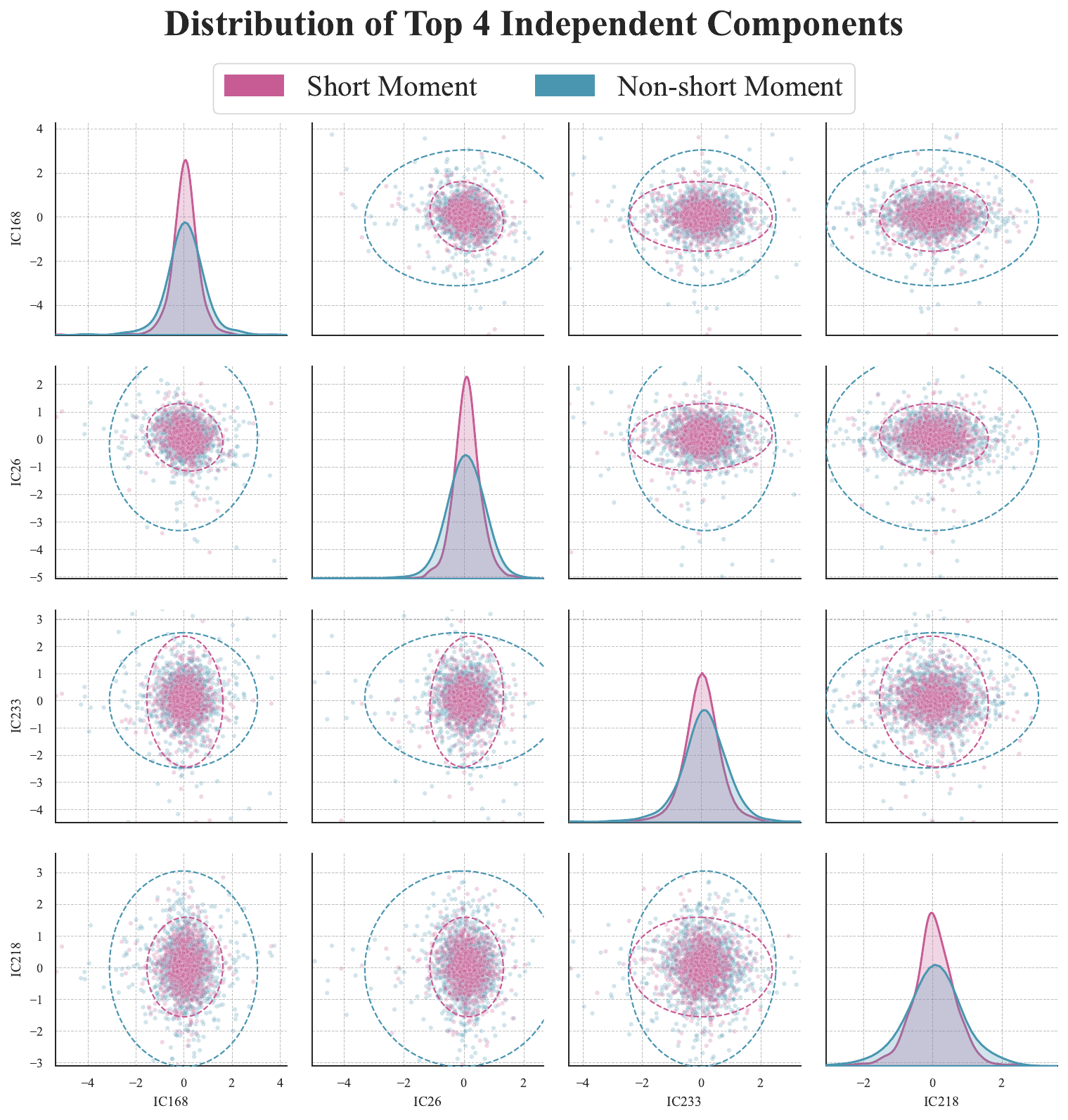}
    \vspace{-1.5mm}
    \caption{
        Pairwise scatter plots for the selected ICs. Diagonals show the 1D density (KDE), while off-diagonals show 2D distributions. The larger 95\% covariance ellipses for the Non-short Moment group provide strong evidence of its greater feature diversity.
    }
    \vspace{-1mm}
    
    \label{fig:supple_aug_feat}
\end{figure}

\noindent \textbf{ICA-Based Feature Diversity Analysis.}
To investigate the feature distributions of the \textsc{QVHighlights} \textit{val} set, we project the features into a latent space constructed using PCA and FastICA. To ensure robustness, we filter out unstable components and retain only those showing high consistency across multiple initialization runs. \Cref{fig:supple_aug_feat} presents the pairwise distributions of the components with the most distinct differences between short and non-short groups. The plot qualitatively demonstrates that non-short moments occupy a broader region in the latent space than short moments, as evidenced by wider kernel density estimates (KDEs) and larger covariance ellipses.

\begin{figure}[h]
    \centering
    \includegraphics[width=0.9\linewidth]{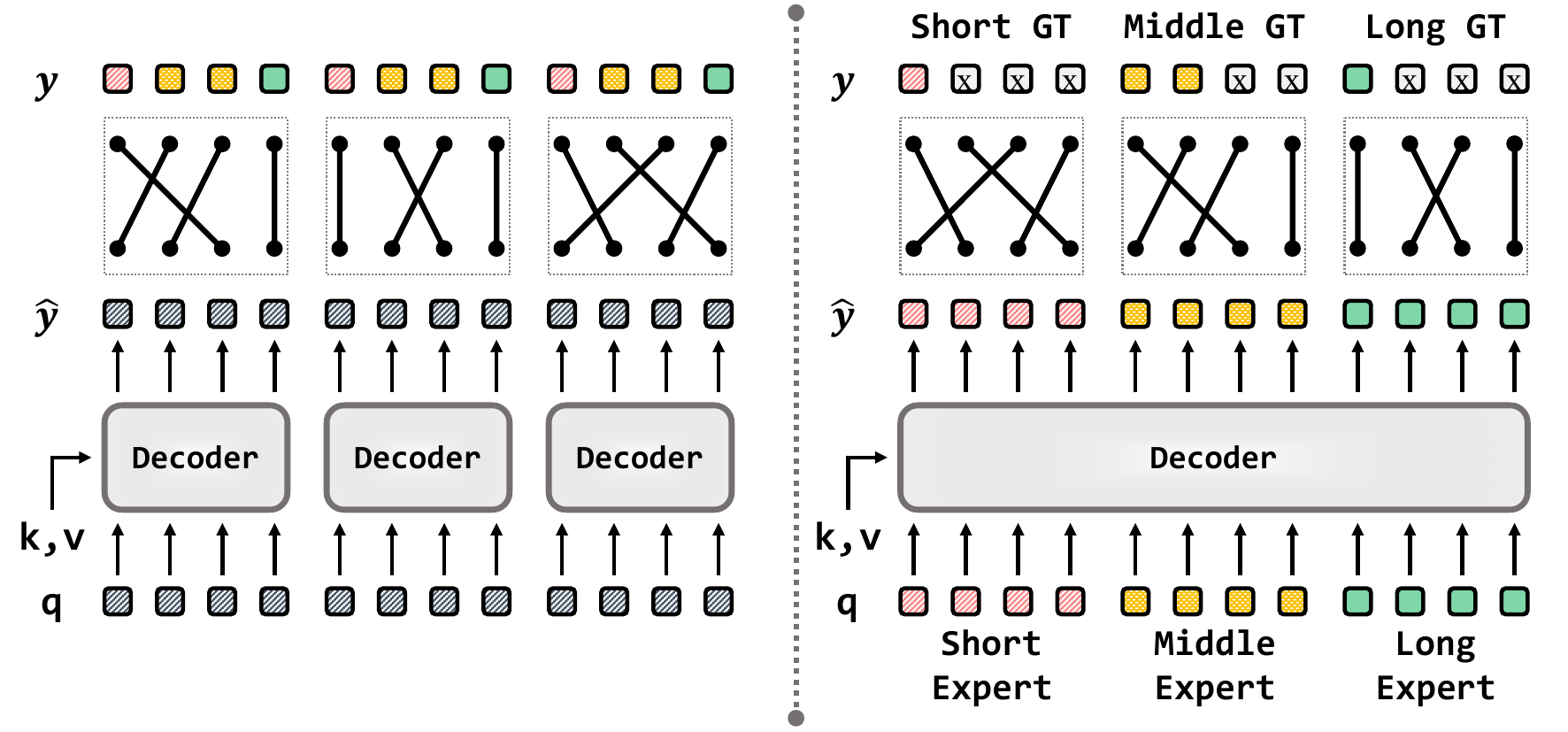}
    \caption{
    [Left] Group-DETR~\cite{chen2023group_Group_DETR} employs one-to-many matching, where the same labels are utilized across all groups.
    [Right] Our length-wise matching is one-to-one, and it operates within each length class.
    By matching only the predictions and ground truths that belong to the same class, this approach enables the creation of length-wise expert queries.
    }
    \vspace{-3mm}
    \label{fig:supple_group}
\end{figure}

\noindent\textbf{Effect of Length-Aware Decoder.} 
We introduce the Length-Aware Decoder (LAD), a novel framework designed to improve moment center predictions by conditioning on moment length. LAD’s core innovation is a length-wise bipartite matching scheme, which assigns queries to predefined length classes to generate length-expert queries. While LAD builds upon Anchor-DETR, Anchor-DETR’s pattern embeddings alone are not length-aware. Furthermore, LAD's strategy differs from the \textit{length class-wise one-to-one} matching of Group-DETR by employing a more precise \textit{length class-wise one-to-one} matching tailored for moment retrieval, as shown in \cref{fig:supple_group}.

The effectiveness of our approach is validated in \Cref{tab:supple_lad} with two key findings. First, applying LAD’s length-wise matching to the Anchor-DETR baseline yields significant gains in both overall and short-moment mAP, proving the value of explicit length conditioning. Second, in contrast to Group-DETR, which improves R@1 but suffer from  significant drop in mAP, LAD achieves substantial improvements in both metrics. These results underscore that LAD's specialized mechanism effectively addresses performance degradation in challenging short moments, establishing it as a more robust and task-specific solution.


\begin{figure*}[!b]
    \centering
 \begin{subfigure}{0.32\textwidth}
        \includegraphics[width=\textwidth]{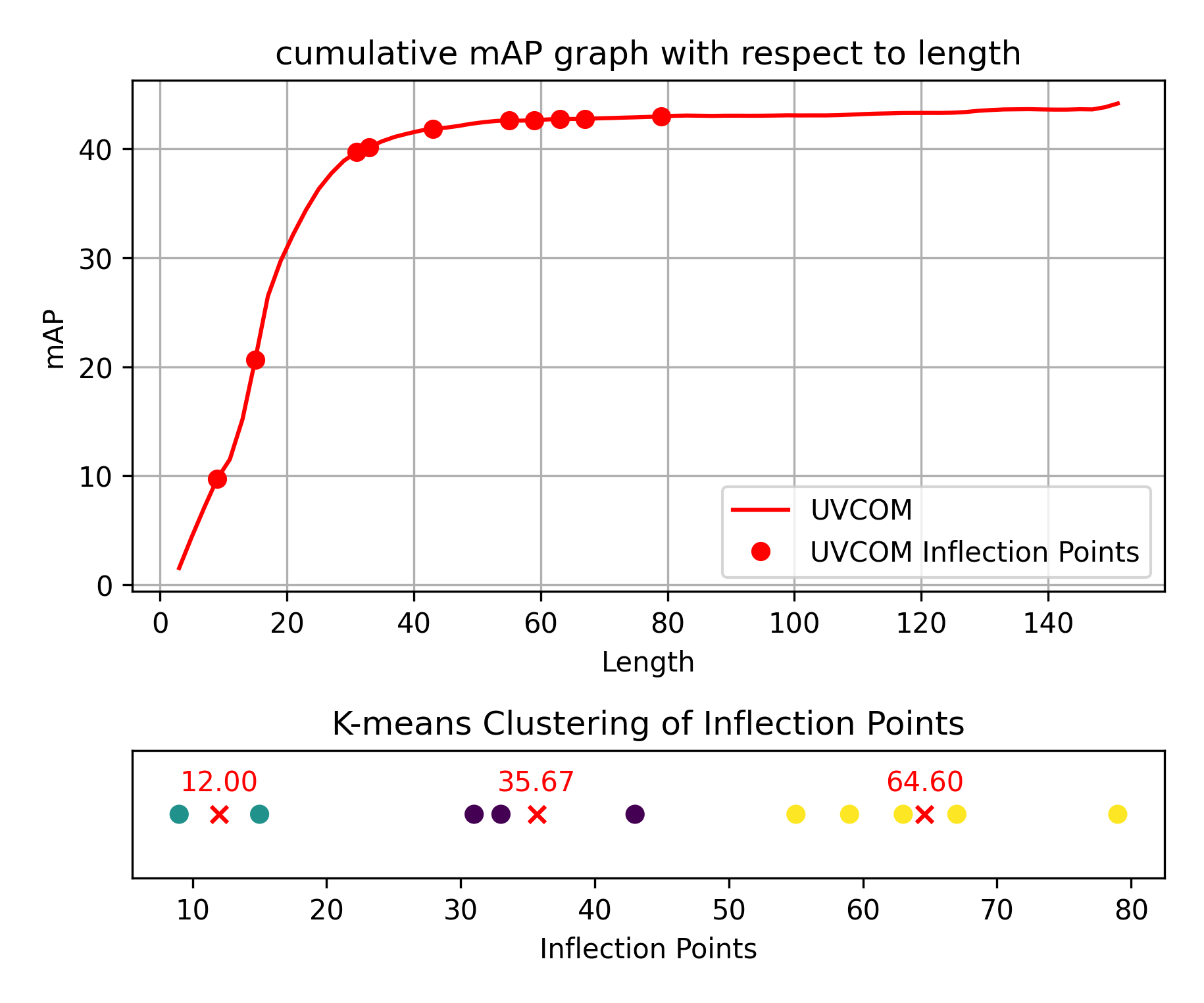}
         \caption{\textsc{QVHighlights}}
    \label{fig:supple_cls_def_qv}
    \end{subfigure}
    \begin{subfigure}{0.32\textwidth}
        \includegraphics[width=\textwidth]{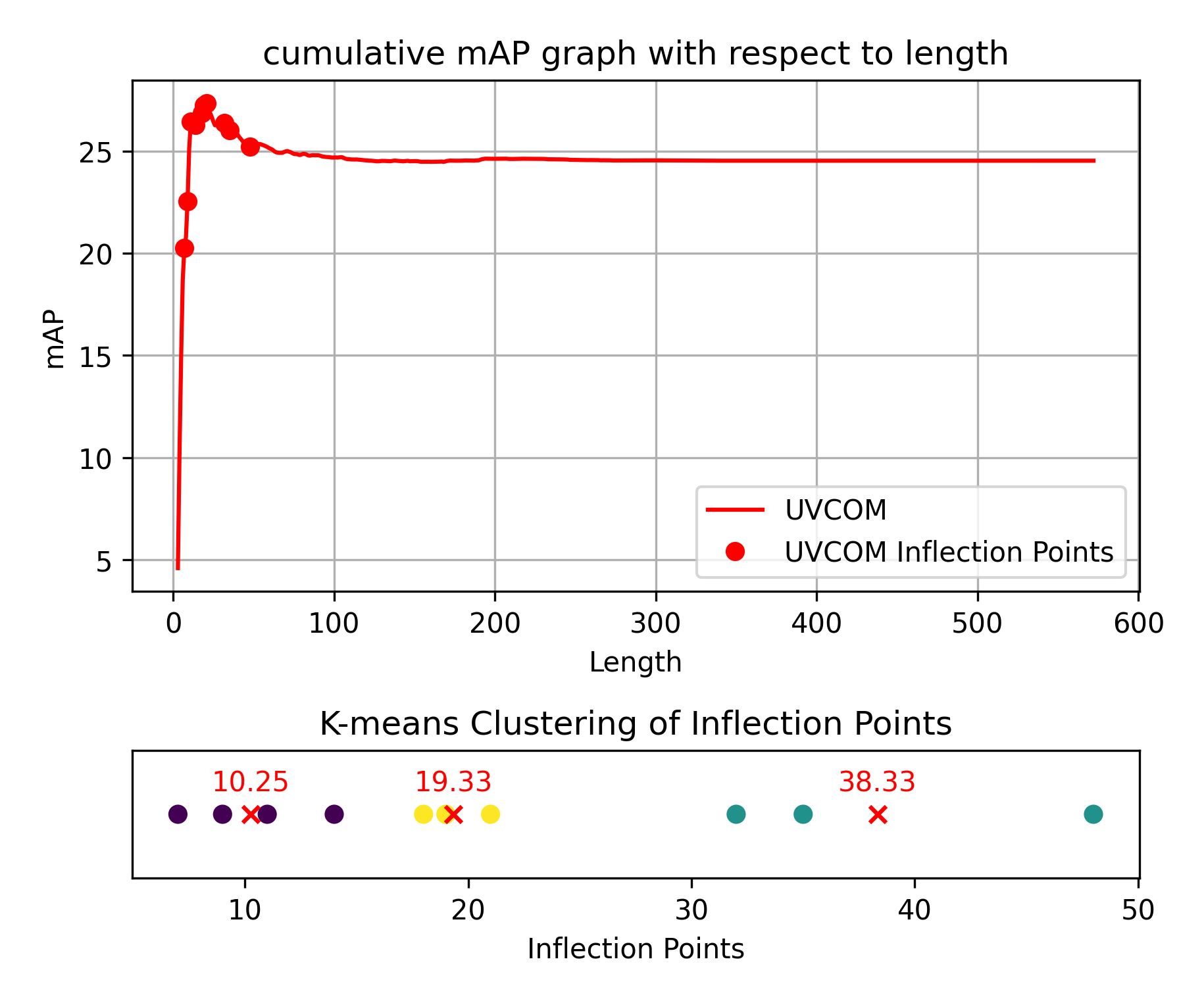}
    \caption{\textsc{TACoS}}
    \label{fig:supple_cls_def_tacos}
    \end{subfigure}
    \begin{subfigure}{0.32\textwidth}
        \includegraphics[width=\textwidth]{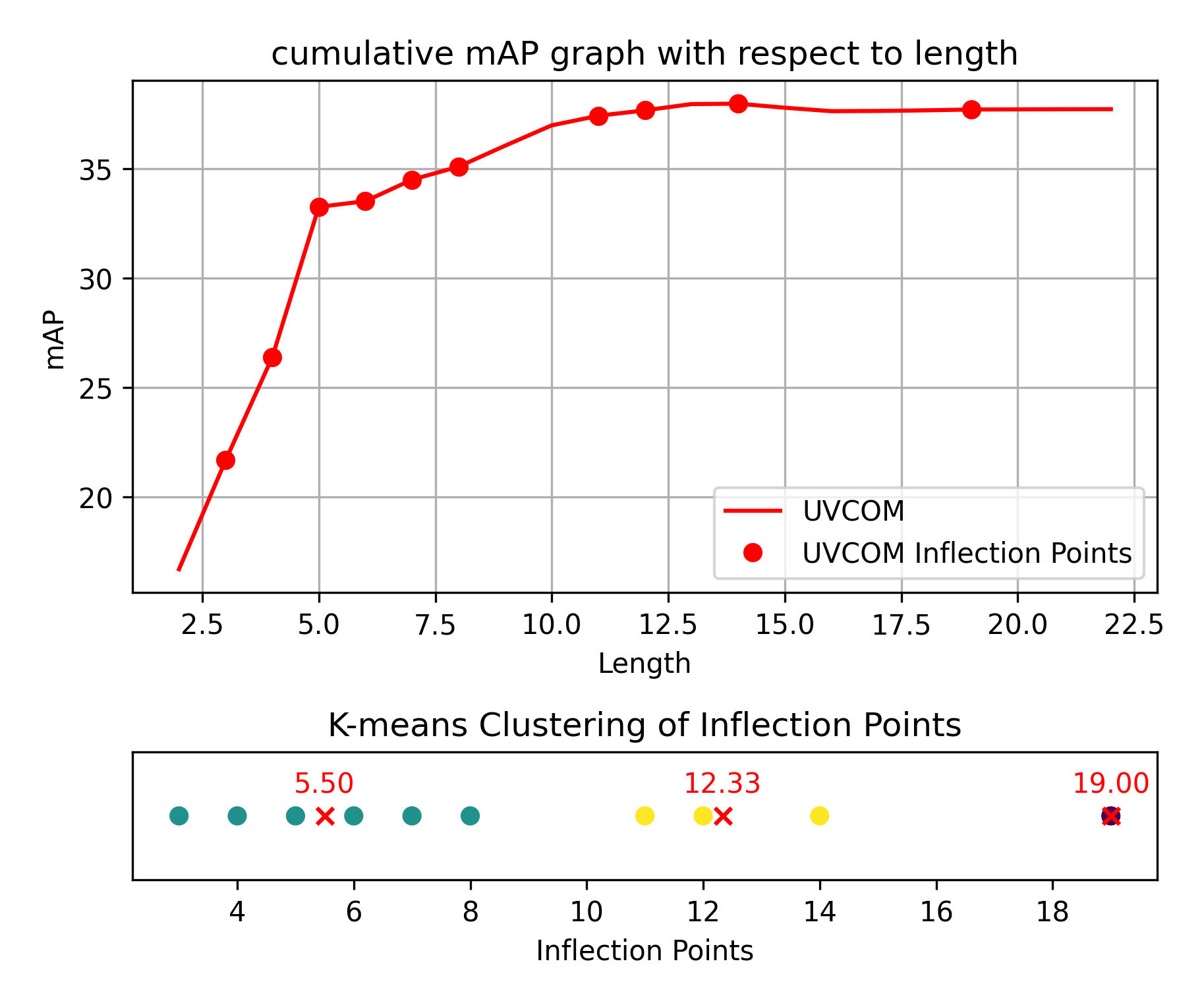}
    \caption{\textsc{Charades-STA}}
    \label{fig:supple_cls_def_cha}
    \end{subfigure}

\caption{We defined length class based on inflection points in the cumulative mAP graph with respect to length.}
\label{fig:supple_cls_def}
\end{figure*}


\input{tables/supple_eff_lad}


\noindent\textbf{Effect of number of queries.} Our method employs 40 queries, with 10 queries allocated to each length class. In comparison, QD-DETR and TR-DETR originally use 10 queries, while UVCOM uses 30. To ensure that the observed performance improvements with our method are not simply due to the increased number of queries, we re-trained the baselines with 40 queries for a fair comparison.

\noindent The results, presented in \Cref{tab:supple_query_num}, clearly demonstrate that increasing the number of queries in the baselines does not guarantee performance gains and can even lead to performance drops, as observed with TR-DETR. This indicates that the performance gains achieved by our method are not trivial or merely due to an increased number of queries.

\input{tables/supple_query_num}


\vspace{0.2cm}
\noindent\textbf{Effect of cut criteria in ForegroundMix.}
We propose ForegroundMix, which cuts a long foreground into shorter sub-foregrounds, shuffles them, and generates new short-moment data. We analyze the effect of $\varepsilon_\text{cut}$, which determines sub-foreground shortening relative to the original long foreground, with QD-DETR as the baseline.

\noindent As shown in \Cref{tab:supple_fgmix_hp}, smaller values of $\varepsilon_\text{cut}$ (more aggressive cutting and greater shortening) lead to improved performance on shorter moments. Regardless of the value, $\varepsilon_\text{cut}$ consistently enhances overall performance. 
Since our primary objective is to improve short-moment performance, we adopt the smallest value, $\varepsilon_\text{cut}=5$, as our default setting.

\input{tables/supple_fgmix_hp}

\vspace{-1.5mm}

\section{Moment Length Class Selection}
\label{sec:supple_cls_def}

\noindent\textbf{Defining length class.}
To define multiple length classes, we select corresponding length thresholds using the cumulative mAP graph with respect to length, as shown in \Cref{fig:supple_cls_def}. We chose cumulative mAP because it effectively highlights lengths where the model underperforms.  
Initially, we compute the cumulative mAP for each moment length based on an existing moment retrieval baseline, UVCOM.
Subsequently, we identify the inflection points on the graph and cluster them using K-means. These clustered points determine the length class thresholds.

\input{tables/supple_audio}

\input{tables/supple_internvideo2}

\vspace{1mm}
\noindent\textbf{Performance comparison based on the number of classes.} The number of classes, $\mathcal{N}_c$, is determined by the value of $k$ in K-means. 
To determine the optimal $k$, we experimented with different class numbers with QD-DETR as the baseline. As shown in \Cref{tab:supple_cls_num}, using four classes resulted in the highest length-awareness in the model.

\vspace{1mm}

\noindent\textbf{Comparison of class-division strategies.} 
We evaluate three substitutes for K-means—equal-interval, equal-count, and Fisher–Jenks variance minimization. As depicted in \cref{tab:supple_length_agnostic}, every alternative yields substantial gains over the baseline, confirming robustness to how length classes are defined. Across all datasets, K-means consistently achieved the best performance, so we adopt it as the default.

\input{tables/supple_cls_num}

\input{tables/supple_length_agonstic}


\section{Evaluation with Diverse Feature Types} \label{sec:supple_audio}

We conducted experiments using various feature types to demonstrate that our methods—MomentMix augmentation and the Length-Aware Decoder—are robust and not limited to specific features.

\vspace{2mm}
\noindent\textbf{Evaluation with additional audio features.} Following prior work, we incorporated additional audio features extracted from PANNs \cite{kong2020panns_panns} to evaluate our method's performance. As shown in \Cref{tab:supple_audio}, compared to the baseline UVCOM trained with the additional audio modality, our method significantly outperforms the baseline, indicating its effectiveness.

\vspace{2mm}
\noindent\textbf{Evaluation with InternVideo2 features.} To further validate the robustness of our method across different feature types, we utilized features from InternVideo2~\cite{wang2024internvideo2}, a recent foundational model for multimodal video understanding, for both video and text modalities. We re-trained the baseline UVCOM and our method using these richer and more powerful features. As shown in \Cref{tab:supple_internvideo2}, despite the enhanced feature quality, the baseline still suffers from performance degradation in short moments. In contrast, our method significantly improves short-moment performance, achieving gains of 9.19\% in R1 and 8.66\% in mAP, along with overall performance improvements. These results demonstrate that our method effectively addresses the short-moment performance issues.



\section{More Qualitative Results}
\label{sec:supple_qualitative}

We provide comparisons with other models across a broader range of samples. Predictions with confidence scores exceeding 0.7 are visualized with alpha = 0.5 on the \textsc{QVHighlights} \textit{val} set; “Ours” denotes UVCOM augmented with our method.

\begin{figure}[ht]
    \centering
    \includegraphics[width=\linewidth]{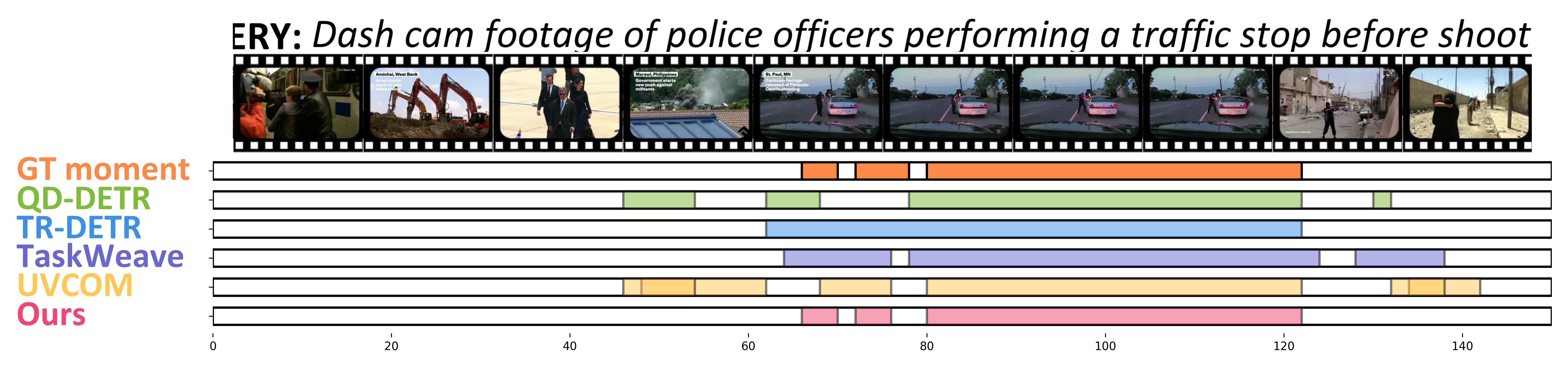}
    \includegraphics[width=\linewidth]{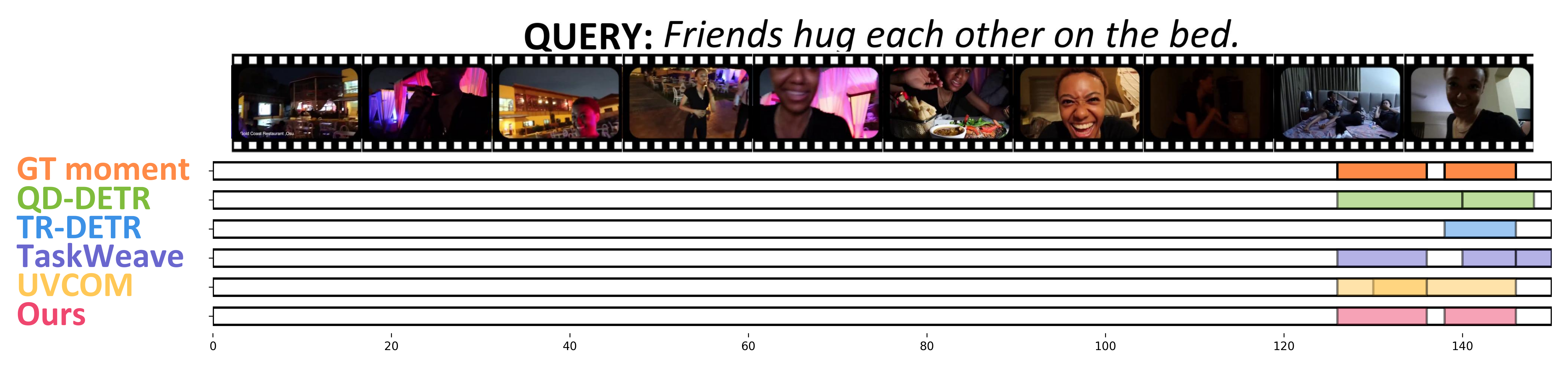}
    \includegraphics[width=\linewidth]{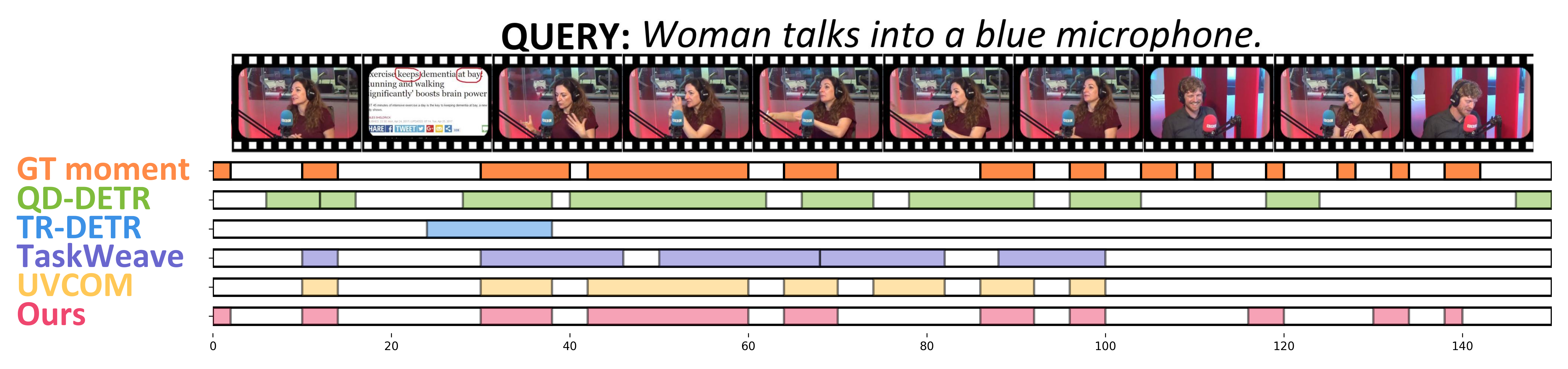}
    \caption{
        Qualitative success on short moments with sharper boundaries and fewer spurious hits.
    }
    \label{fig:supple_qualitative}
\end{figure}

As shown in \Cref{fig:supple_qualitative}, Existing models frequently struggle to effectively distinguish between foreground and background, resulting in inaccurate predictions or missed detections of short moments. In contrast, our model excels in accurately and robustly predicting short moments. our approach yields consistently higher accuracy on short moments.

\begin{figure}[ht]
    \centering
    \includegraphics[width=\linewidth]{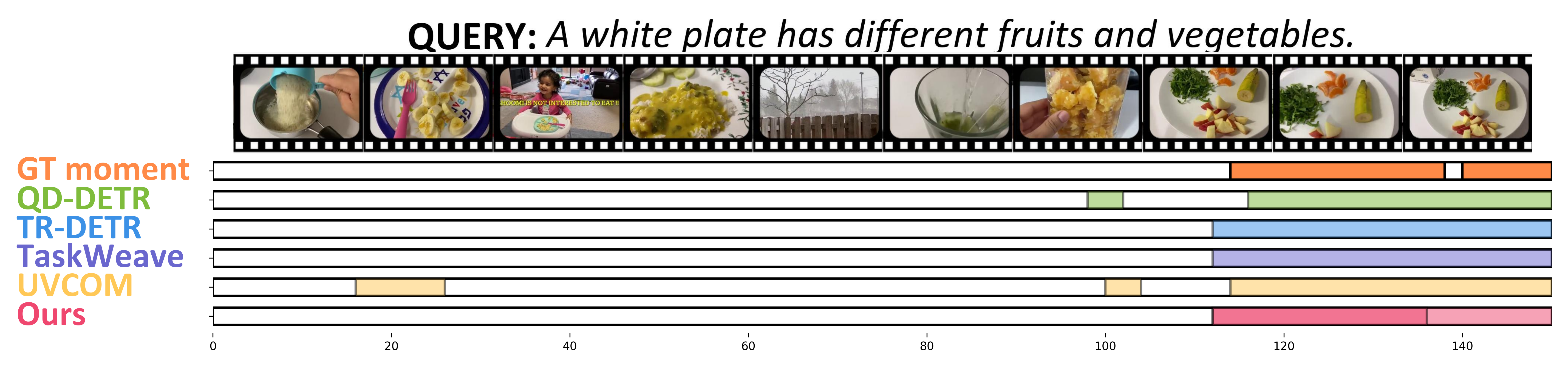}
    \includegraphics[width=\linewidth]{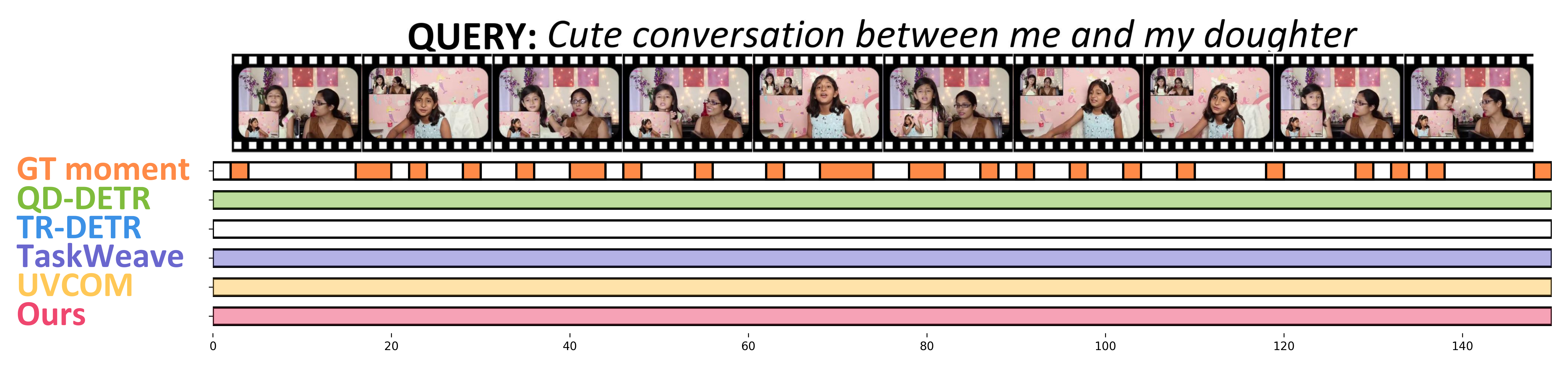}
    \caption{
        Qualitative failure driven by difficult visual–text alignment.
    }
    \label{fig:supple_qualitative_2}
\end{figure}

We also include representative failure cases that are difficult irrespective of length, primarily due to visual–text alignment challenges. Examples are strong window reflections that obscure content and screen-in-screen scenes that induce visual ambiguity as shown in \Cref{fig:supple_qualitative_2}.

\stopcontents[supplement]

%% file: tables/supple_eff_lad.tex

\begin{table}[ht]

\setlength{\tabcolsep}{0.7em}
\centering

\vspace{-1mm}

\begin{small}
\setlength{\tabcolsep}{2pt}

\resizebox{0.90\linewidth}{!}
{

    \begin{tabular}{
        l    c c   c c   c c    c c
    }

    \toprule

    \multirow{3}{*}{Method} & \multicolumn{2}{c}{{Short}} & \multicolumn{2}{c}{Middle} & \multicolumn{2}{c}{Long}  & \multicolumn{2}{c}{Full} \\

    \cmidrule(l){2-3}  \cmidrule(l){4-5}  \cmidrule(l){6-7}  \cmidrule(l){8-9}

      & R1 & mAP & R1 & mAP & R1 & mAP  & R1 & mAP \\

    \midrule


Baseline	 & 4.57	 & 7.77	 & 38.89	 & 43.10	 & 42.62	 & 47.44	 & 41.06	 & 41.00	\\

\hline

\( \)Anchor-DETR	  & 4.48	 & 7.43	 & 39.53	 & 43.36	 & 45.4	 & 49.09	 & 42.46	 & 41.55	\\

\( \)Group-DETR	  & 4.90	 & 3.84	 & {40.24}	 & 40.72	 & 43.14	 & 43.79	 & 42.17	 & 37.97	\\

\rowcolor{olive!10}~LAD	 & \textbf{8.76}	 & \textbf{11.01}	 & \textbf{40.55}	 & \textbf{45.53}	 & \textbf{43.69}	 & \textbf{50.76}	 & \textbf{43.65}	 & \textbf{44.48}	\\

    \bottomrule
    \end{tabular}
}

\end{small}

\caption{Performance comparison on \textsc{QVHighlights} \textit{val} set.}

\label{tab:supple_lad}

\end{table}

%% file: tables/supple_query_num.tex

\begin{table}[ht]

\setlength{\tabcolsep}{0.3em}
\centering

\begin{small}
\setlength{\tabcolsep}{2pt}

\resizebox{0.95\linewidth}{!}
{

    \begin{tabular}{
        l    c c   c c   c c    c c
    }

    \toprule

    \multirow{3}{*}{Method} & \multicolumn{2}{c}{{Short}} & \multicolumn{2}{c}{Middle} & \multicolumn{2}{c}{Long}  & \multicolumn{2}{c}{Full} \\

    \cmidrule(l){2-3}  \cmidrule(l){4-5}  \cmidrule(l){6-7}  \cmidrule(l){8-9}

      & R1 & mAP & R1 & mAP & R1 & mAP  & R1 & mAP \\

    \midrule

QD-DETR	 & 4.45	 & 8.34	 & 39.54	 & 43.54	 & 43.89	 & 47.80	 & 41.90	 & 41.24	\\
QD-DETR$^{\dag}$	 & 5.48	 & 8.88	 & 40.17	 & 43.95	 & 40.70	 & 44.17	 & 41.39	 & 40.29	\\
\rowcolor{olive!10}QD-DETR+Ours	  & \textbf{11.07}	 & \textbf{15.27}	 & \textbf{43.12}	 & \textbf{48.53}	 & \textbf{44.39}	 & \textbf{52.65}	 & \textbf{46.13}	 & \textbf{47.70}	\\

\hline

TR-DETR	 & 5.80	 & 9.91	 & 44.01	 & 46.95	 & 47.35	 & 51.70	 & 46.32	 & 45.10	\\
TR-DETR$^{\dag}$	 & 3.66	 & 7.32	 & 39.44	 & 43.01	 & \textbf{47.39}	 & 50.07	 & 42.91	 & 41.83	\\
\rowcolor{olive!10}TR-DETR+Ours	  & \textbf{9.91}	 & \textbf{15.17}	 & \textbf{47.11}	 & \textbf{51.83}	 & 46.78	 & \textbf{53.53}	 & \textbf{49.15}	 & \textbf{49.80}	\\

\hline

UVCOM	 & 5.97	 & 12.65	 & 45.97	 & 49.04	 & 45.19	 & 49.39	 & 46.77	 & 45.80	\\
UVCOM$^{\dag}$	 & 5.48	 & 10.39	 & 40.17	 & 47.82	 & 40.70	 & 47.79	 & 41.39	 & 43.87	\\
\rowcolor{olive!10}UVCOM+Ours	  & \textbf{12.80}	 & \textbf{18.46}	 & \textbf{46.87}	 & \textbf{52.36}	 & \textbf{46.92}	 & \textbf{53.23}	 & \textbf{49.85}	 & \textbf{50.76}	\\

    \bottomrule
    \end{tabular}
}

\end{small}

\caption{Results on \textsc{QVHighlights} \textit{val} set. {\dag} indicates training with the number of queries the same as ours.}

\label{tab:supple_query_num}

\end{table}

%% file: tables/supple_fgmix_hp.tex
\begin{table}[ht]

\setlength{\tabcolsep}{0.3em}
\centering

\begin{small}
\setlength{\tabcolsep}{2pt}
\vspace{-2mm}
\resizebox{0.90\linewidth}{!}
{

    \begin{tabular}{
        l    c c   c c   c c    c c
    }

    \toprule

    \multirow{3}{*}{Method} & \multicolumn{2}{c}{\textbf{Short}} & \multicolumn{2}{c}{Middle} & \multicolumn{2}{c}{Long}  & \multicolumn{2}{c}{Full} \\

    \cmidrule(l){2-3}  \cmidrule(l){4-5}  \cmidrule(l){6-7}  \cmidrule(l){8-9}

      & R1 & mAP & R1 & mAP & R1 & mAP  & R1 & mAP \\    

    \midrule


Baseline\( \quad\quad \)	 & 4.57	 & 7.77	 & 38.89	 & 43.10	 & 42.62	 & 47.44	 & 41.06	 & 41.00	\\
\hline
\rowcolor{gray!10} $\varepsilon_\text{cut}=5$	 & 7.86	 & 12.21	 & 41.42	 & 45.28	 & 43.45	 & 47.69	 & 43.84	 & 43.32	\\
$\varepsilon_\text{cut}=10$	 & 6.78	 & 10.87	 & 42.31	 & 46.07	 & 44.16	 & 48.11	 & 44.35	 & 43.45	\\
$\varepsilon_\text{cut}=15$	 & 5.45	 & 8.68	 & 41.35	 & 44.78	 & 44.34	 & 48.37	 & 43.46	 & 42.48	\\

    \bottomrule
    \end{tabular}
    
}

\end{small}

\caption{Performance comparison on \textsc{QVHighlights} \textit{val} set. $\varepsilon_\text{cut}$ controls sub-foreground shortening in ForegroundMix. Across all values, $\varepsilon_\text{cut}$ consistently improving overall performance, with smaller values excelling in short-moment enhancement. } 
\label{tab:supple_fgmix_hp}

\end{table}

%% file: tables/supple_audio.tex
\begin{table*}[hb]

    \centering
    \vspace{-2mm}

\setlength{\tabcolsep}{7pt}
\resizebox{0.85\linewidth}{!}
{
    \begin{tabular}{
        l   c c   c c   c c   c c c    c c c 
    }
    \toprule

    {\multirow{3}{*}{Method}} & \multicolumn{2}{c}{Short}  & \multicolumn{2}{c}{Middle }  & \multicolumn{2}{c}{Long } & \multicolumn{6}{c}{Full}  \\

    \cmidrule(l){2-3}  \cmidrule(l){4-5} \cmidrule(l){6-7} \cmidrule(l){8-13} 
    
     & R1 & mAP & R1 & mAP & R1 & mAP & \multicolumn{3}{c}{R1} & \multicolumn{3}{c}{mAP}  \\

    \cmidrule(l){2-3}  \cmidrule(l){4-5} \cmidrule(l){6-7} \cmidrule(l){8-10}  \cmidrule(l){11-13} 
    
     &\multicolumn{2}{c}{Avg.} & \multicolumn{2}{c}{Avg.} & \multicolumn{2}{c}{Avg.} & @0.5 & @0.7 & Avg.  & @0.5 & @0.75 & Avg.  \\
    \midrule
    
UVCOM$\ddagger$		 & 4.45	 & 11.00	 & 44.18	 & 48.03	 & 43.89	 & 48.84	 & 64.26	 & 49.42	 & 44.76	 & 64.92	 & 45.29	 & 44.70	\\
\rowcolor{olive!10} ~ + Ours		 & 13.38	 & 17.77	 & 45.51	 & 51.12	 & 45.84	 & 53.51	 & 66.71	 & 52.97	 & 48.77	 & 68.04	 & 51.52	 & 50.10	\\
		\vspace{-4pt} & \gainp{+8.93} 	& \gainp{+6.77} 	& \gainp{+1.33} 	& \gainp{+3.09} 	& \gainp{+1.95} 	& \gainp{+4.67} 	& \gainp{+2.45} 	& \gainp{+3.55} 	& \gainp{+4.01} 	& \gainp{+3.12} 	& \gainp{+6.23} 	& \gainp{+5.40} 	\\

        \bottomrule
    \end{tabular}
}

\caption{Performance comparison on the \textsc{QVHighlights} \textit{val} set using additional audio modality. {\ddag} means the result reproduced from the original repository.
}
\label{tab:supple_audio}

\end{table*}

%% file: tables/supple_internvideo2.tex
\begin{table*}[hb]

    \centering
    \vspace{-2mm}

\setlength{\tabcolsep}{7pt}
\resizebox{0.85\linewidth}{!}
{
    \begin{tabular}{
        l   c c   c c   c c   c c c    c c c 
    }
    \toprule

    {\multirow{3}{*}{Method}} & \multicolumn{2}{c}{Short}  & \multicolumn{2}{c}{Middle }  & \multicolumn{2}{c}{Long } & \multicolumn{6}{c}{Full}  \\

    \cmidrule(l){2-3}  \cmidrule(l){4-5} \cmidrule(l){6-7} \cmidrule(l){8-13} 
    
     & R1 & mAP & R1 & mAP & R1 & mAP & \multicolumn{3}{c}{R1} & \multicolumn{3}{c}{mAP}  \\

    \cmidrule(l){2-3}  \cmidrule(l){4-5} \cmidrule(l){6-7} \cmidrule(l){8-10}  \cmidrule(l){11-13} 
    
     &\multicolumn{2}{c}{Avg.} & \multicolumn{2}{c}{Avg.} & \multicolumn{2}{c}{Avg.} & @0.5 & @0.7 & Avg.  & @0.5 & @0.75 & Avg.  \\
    \midrule
    
UVCOM$\ddagger$		 & 5.64	 & 10.83	 & 48.96	 & 51.12	 & 49.16	 & 51.71	 & 70.13	 & 54.97	 & 49.99	 & 67.95	 & 47.88	 & 47.56	\\
\rowcolor{olive!10} ~ + Ours	 & 14.83	 & 19.49	 & 49.23	 & 54.60	 & 49.56	 & 55.67	 & 71.10	 & 58.00	 & 52.85	 & 71.45	 & 54.84	 & 53.25	\\
	\vspace{-4pt} & \gainp{+9.19} 	& \gainp{+8.66} 	& \gainp{+0.27} 	& \gainp{+3.48} 	& \gainp{+0.40} 	& \gainp{+3.96} 	& \gainp{+0.97} 	& \gainp{+3.03} 	& \gainp{+2.86} 	& \gainp{+3.50} 	& \gainp{+6.96} 	& \gainp{+5.69} 	\\

        \bottomrule
    \end{tabular}
}

\caption{Performance comparison on the \textsc{QVHighlights} \textit{val} set using the $\text{InternVideo2}_{s2}\text{-6B}$ features for both video and text modalities. {\ddag} means the result reproduced from the original repository.
}
\label{tab:supple_internvideo2}
    
\end{table*}

%% file: tables/supple_cls_num.tex
\begin{table}[ht]

\setlength{\tabcolsep}{0.3em}
\centering

\vspace{-2mm}

\begin{small}
\setlength{\tabcolsep}{2pt}

\resizebox{0.90\linewidth}{!}
{

    \begin{tabular}{
        l    c c   c c   c c    c c
    }

    \toprule

    \multirow{3}{*}{Method} & \multicolumn{2}{c}{\textbf{Short}} & \multicolumn{2}{c}{Middle} & \multicolumn{2}{c}{Long}  & \multicolumn{2}{c}{Full} \\

    \cmidrule(l){2-3}  \cmidrule(l){4-5}  \cmidrule(l){6-7}  \cmidrule(l){8-9}

      & R1 & mAP & R1 & mAP & R1 & mAP  & R1 & mAP \\    

    \midrule

Baseline\( \quad\quad \) 	 & 4.57	 & 7.77	 & 38.89	 & 43.10	 & 42.62	 & 47.44	 & 41.06	 & 41.00	\\
\hline
$\mathcal{N}_c = 2$	 & 6.95	 & 9.01	 & 38.19	 & 44.95	 & 43.36	 & 49.80	 & 41.56	 & 42.99	\\
$\mathcal{N}_c = 3$	 & 7.20	 & 9.64	 & 38.54	 & 44.70	 & 41.29	 & 49.13	 & 41.08	 & 43.03	\\
\rowcolor{gray!10}$\mathcal{N}_c = 4$  &  {8.76}	 & {11.01}	 & {40.55}	 & {45.53}	 & {43.69}	 & {50.76}	 & {43.65}	 & {44.48}	\\

    \bottomrule
    \end{tabular}
}

\end{small}

\caption{Performance comparison on \textsc{QVHighlights} \textit{val} set. $\mathcal{N}_c$ indicates the number of length classes in LAD.} 
\label{tab:supple_cls_num}

\end{table}

%% file: tables/supple_length_agonstic.tex

\begin{table}[ht]

\setlength{\tabcolsep}{0.3em}
\centering

\vspace{-2mm}

\begin{small}
\setlength{\tabcolsep}{2pt}

\resizebox{0.90\linewidth}{!}
{

    \begin{tabular}{
        l    c c   c c   c c    
    }

    \toprule

    \multirow{3}{*}{Method} & \multicolumn{2}{c}{\textsc{QVHighlights}} & \multicolumn{2}{c}{\textsc{TACoS}} & \multicolumn{2}{c}{\textsc{Charades-STA}}   \\

    \cmidrule(l){2-3}  \cmidrule(l){4-5}  \cmidrule(l){6-7}  

      & R1 avg. & mAP avg. & R1@0.5 & R1@0.7 & R1@0.5 & R1@0.7  \\

    \midrule


UVCOM	 & 46.77	 & 45.80	 & 36.39	 & 23.32	 & 59.25	 & 36.64	\\
\hline
\rowcolor{gray!10}K-means	 & 49.85	 & 50.76	 & 42.21	 & 28.02	 & 61.45	 & 40.22	\\
Equal-Interval	 & 48.86	 & 49.21	 & 41.41	 & 27.99	 & 60.67	 & 40.51	\\
Equal-Count	 & 48.39	 & 50.42	 & 41.54	 & 27.09	 & 58.47	 & 38.66	\\
Fisher-Jenks	 & 47.65	 & 49.58	 & 41.01	 & 26.84	 & 60.19	 & 40.86	\\
        
    \bottomrule
    \end{tabular}
}

\end{small}

\caption{Performance Comparison of different strategies for moment length class definition. Across all datasets, all strategies improve performance over the baseline; K-means yields the best results overall.} 

\label{tab:supple_length_agnostic}

\end{table}